\documentclass{article}

\PassOptionsToPackage{numbers, compress}{natbib}

\usepackage[preprint]{neurips_2026}

\usepackage[utf8]{inputenc}
\usepackage[T1]{fontenc}
\usepackage{hyperref}
\usepackage{url}
\usepackage{booktabs}
\usepackage{amsfonts}
\usepackage{amsmath}
\usepackage{nicefrac}
\usepackage[nopatch=footnote]{microtype}
\usepackage{xcolor}
\usepackage{graphicx}
\graphicspath{{./}{../}}
\usepackage{subcaption}
\usepackage{wrapfig}

\usepackage{pifont}
\usepackage{xspace}

\usepackage{multirow}
\usepackage{algorithm}
\usepackage{algpseudocode}
\usepackage{xspace}
\usepackage{float}

\newcommand{\best}[1]{\textbf{#1}}
\newcommand{\second}[1]{\underline{#1}}
\newcommand{\third}[1]{#1}
\newcommand{\offmark}{$^{\dagger}$}
\newcommand{\modelname}{\mbox{\textcolor{black}{$R^3$}}\xspace}

\newcommand{\modelstream}{\mbox{\textcolor{black}{$R^3$-Stream}}\xspace}

\newcommand{\modelnamebf}{\mbox{\textcolor{black}{$\mathbf{R}^3$}}\xspace}

\newcommand{\figref}[1]{Fig.~\ref{#1}}
\newcommand{\secref}[1]{Sec.~\ref{#1}}

\newcommand{\tabref}[1]{Tab.~\ref{#1}}
\newcommand{\appref}[1]{App.~\ref{#1}}

\title{$R^3$: 3D Reconstruction via Relative Regression}
\author{
  Congrong Xu\textsuperscript{1,2}\thanks{Research done during an internship at Westlake University.} \quad
  Huachen Gao\textsuperscript{2} \quad
  Xingyu Chen\textsuperscript{2} \quad
  Yuliang Xiu\textsuperscript{2} \quad
  Jun Gao\textsuperscript{1,3} \quad
  Anpei Chen\textsuperscript{2}\thanks{Corresponding author.} \\
  \rule{0pt}{16pt}\textsuperscript{1}University of Michigan \quad
  \textsuperscript{2}Westlake University \quad\textsuperscript{3}NVIDIA
}

\begin{document}

\maketitle

\begin{abstract}
Recent feed-forward geometry foundation models have demonstrated impressive generalization by recovering depth and poses in a single forward pass. However, these models are typically constrained by a global coordinate frame assumption. This dependency becomes a significant bottleneck for long-context and streaming reconstruction, as it forces the network to maintain an arbitrary temporal origin and handle translation magnitudes that grow unbounded over time. Our solution, which we call $R^3$, employs relative regression. We employ a lightweight MLP to predict confidence-weighted relative constraints. These confidences serve as a unified anchor: weighting losses during training and guiding pose aggregation during inference. $R^3$ supports both full-context offline reconstruction and causal, bounded-memory streaming. Our evaluation in both offline and streaming settings validates the effectiveness of our relative mechanism. Project Page: \href{https://kevinxu02.github.io/r3-site/}{\textcolor{magenta}{\tt\textit{kevinxu02.github.io/r3-site}}}
\end{abstract}

\section{Introduction}\label{sec:introduction}
\suppressfloats[t]
\setcounter{topnumber}{1}

\begin{figure}[t]
  \vspace{-20pt}
  \centering
  \includegraphics[width=\linewidth]{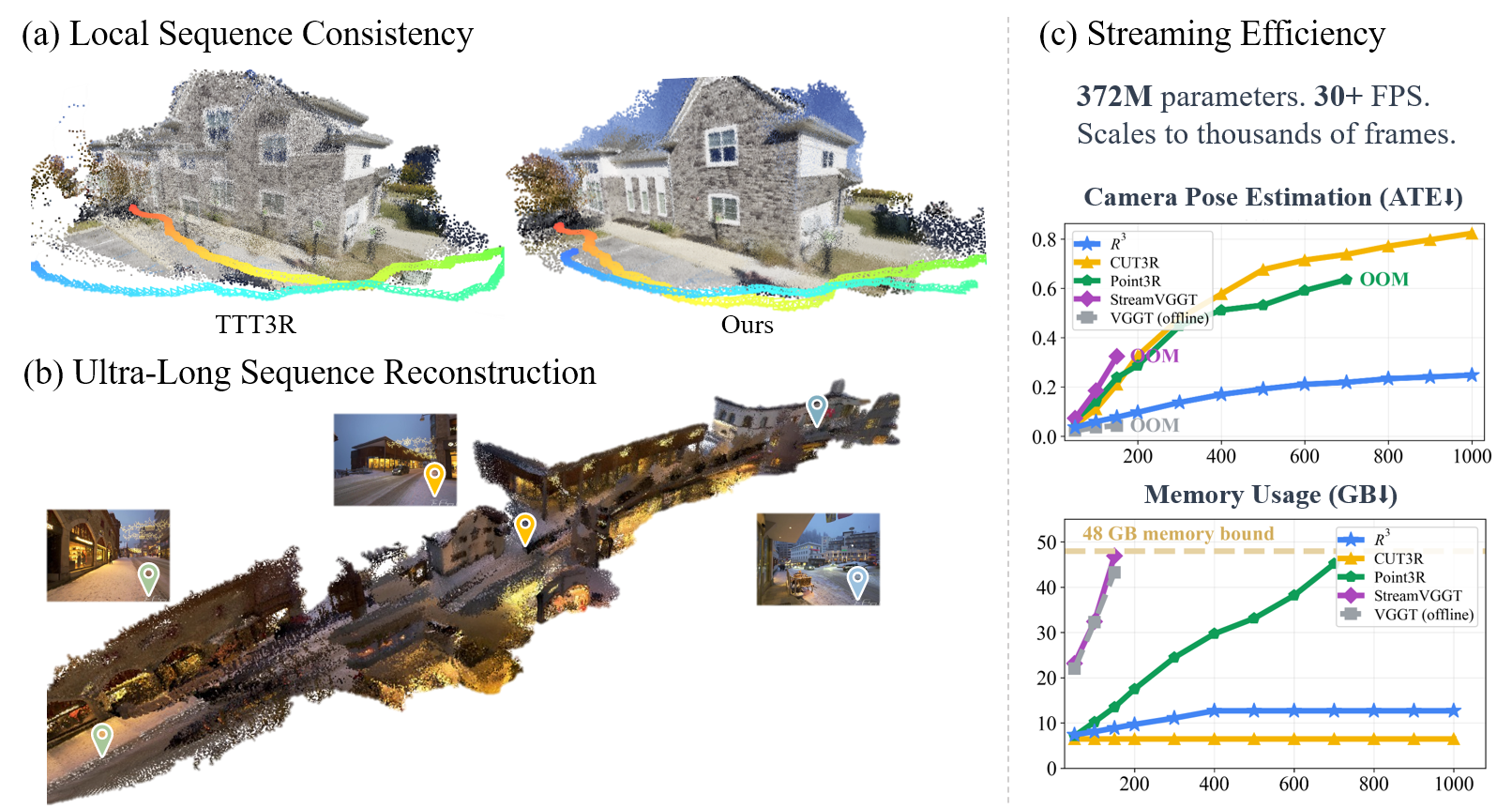}
  \caption{\textbf{Consistent, scalable, and efficient streaming geometry via relative pose regression.}
\modelname reconstructs camera poses and dense geometry from unbounded video streams via feed-forward relative pose regression.
It maintains local consistency, scales to ultra-long sequences with bounded memory, and runs at 20+ FPS with 372M parameters.}
  \label{fig:teaser}
  \vspace{-12pt}
\end{figure}

Feed-forward geometry foundation models have made camera and geometry prediction much more robust. Systems such as DUSt3R, VGGT, $\pi^3$, and DA3~\cite{wang2024dust3r,wang2025vggt,wang2026pi3,lin2025da3} are able to recover depth, poses, or pointmaps from images in one forward pass. However, most of these models are built around a global coordinate frame: they either predict pointmaps in a shared frame or regress every camera pose relative to one sequence-level world frame, such as the first camera or a learned canonical frame. This design works well for short offline clips, but it becomes problematic for long videos and streaming reconstruction, where the model must keep a consistent global frame as new views arrive.

Recent work has begun to relax this global-frame assumption. In particular, $\pi^3$~\cite{wang2026pi3} replaces the fixed absolute pose loss with relative pose supervision, which reduces the bias toward a hard-coded temporal origin. Yet the output is still a set of global poses in one coordinate system. The model can choose the origin more flexibly, but it still has to represent an entire trajectory in a single frame, and translation magnitudes can grow as the sequence becomes longer. For online reconstruction, this keeps a difficult coordinate-choice problem inside the neural network: future poses are heavily biased toward the frame coordinate initialized from earlier observations.

We suggest that feed-forward 3D reconstruction should learn local relations between views and assemble the global trajectory afterward. Relative pose is a better target for long-context and streaming settings for two simple reasons: its scale depends on the baseline between two frames rather than on the total trajectory length, which keeps regression in distribution as sequences grow; and an $N$-frame sequence yields many pairwise constraints instead of only $N$ absolute poses. However, raw pairwise prediction is not sufficient: pairs differ in visibility, texture, motion, and baseline, and the model must regress all of them, and global assembly provides no signal for which predictions to trust. To track this issue, we incorporate a confidence modeling for each relative pose, allowing the model to emphasize reliable pairs. This enables any registered frame to serve as an anchor for a new one, with the system selecting the most confident reference from the registered set instead of chaining all poses to a fixed origin.

\begin{figure}[t]
  \vspace{-20pt}
  \centering
  \includegraphics[width=\linewidth]{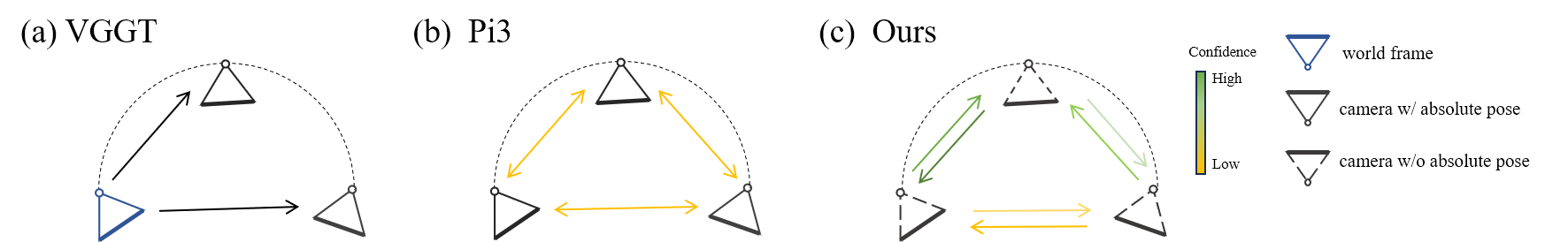}
  \caption{\textbf{Three feed-forward pose paradigms, viewed as pose graphs.}  Edges denote supervised pairwise pose terms; arrowheads encode directional supervision. (a) VGGT~\cite{wang2025vggt} fixes the world frame to the first camera and supervises only edges from this anchor to every other camera. (b) $\pi^3$~\cite{wang2026pi3} regresses absolute poses in a model-chosen world frame and supervises every unordered pair with uniform weight. (c) \modelname drops the global-pose head and supervises every directed pair $(i,j)$ with a learned per-edge confidence, yielding a fully-connected directed pose graph.}
  \label{fig:method_comparison}
  \vspace{-15pt}
\end{figure}

More specifically, we introduce 3D Reconstruction via Relative Regression (\textit{\modelname} for short), a feed-forward reconstruction framework based on pairwise relative pose. \figref{fig:method_comparison} contrasts our supervision target with VGGT's first-frame-anchored absolute poses and $\pi^3$'s model-chosen world frame: \modelname does not ask the network to regress global poses directly, but instead assembles them downstream from pairwise predictions with learned per-pair confidence. Using camera tokens from a DA3 backbone, a lightweight MLP predicts relative rotation, relative translation, and separate confidence scores for rotation and translation. These confidences are used during training as reliability weights and during inference as aggregation weights that fuse pairwise predictions into a consistent trajectory. By focusing on local relations, the model treats absolute poses as a downstream assembly task.

\modelname is a causal, bounded-memory streaming model. Each incoming frame is placed by aggregating confident relative-pose predictions against a confidence-driven keyframe bank that serves as the active context. For offline use, the same checkpoint can run with the causal mask disabled, providing a full-context inference mode without retraining or a second model.

With only 372M parameters and trained on 6 48GB GPUs, \modelname is roughly a third the size of recent 1B-class feed-forward reconstruction models and uses substantially smaller training resources, yet matches or surpasses them on pose estimation and dense reconstruction across diverse 3D tasks. Compared with state-of-the-art streaming methods~\cite{wang2025cut3r,wu2025point3r,lan2025stream3r,zhuo2026streamvggt,chen2025ttt3r,yuan2026infinitevggt}, it achieves competitive accuracy while preserving bounded memory usage, even for streams containing thousands of frames.

In summary, we make the following contributions:
(i)~We reformulate feed-forward 3D reconstruction as a pairwise relative-pose regression task, reducing direct dependence on a fixed global coordinate frame and enabling confidence-weighted global assembly.
(ii)~We introduce a lightweight MLP that predicts separate rotation and translation confidences, reusing them for loss weighting, streaming pose aggregation, and keyframe-bank management.
(iii)~We present a single causal architecture that supports both bounded-memory streaming and full-context inference via a simple test-time attention switch, achieving high accuracy across short and long sequences.
\setcounter{topnumber}{2}

\section{Related Work}\label{sec:related}

\paragraph{Traditional 3D Reconstruction and Learned Back-ends.}
Classical pipelines split reconstruction into SfM~\cite{schonberger2016structure,pan2024glomap}, MVS~\cite{furukawa2010accurate,schonberger2016pixelwise,yao2018mvsnet}, and SLAM~\cite{klein2007ptam,murartal2017orbslam2,engel2017dso,campos2021orbslam3}, with learning gradually folded in via features and matchers~\cite{detone2018superpoint,sarlin2020superglue,lindenberger2023lightglue,sun2021loftr,wang2024eloftr}, learned SfM/VO/SLAM components~\cite{wang2024vggsfm,duisterhof2025mast3rsfm,teed2021droidslam,teed2023dpvo}, and learned-prior systems that integrate dense 3D predictions with keyframes, pose graphs, sparse volumes, or trajectory smoothing~\cite{murai2025mast3rslam,maggio2025vggtslam,liu2025slam3r,wang2025amb3r,taher2025kvtracker,li2025megasam}.
We follow the learned-prior direction but keep most of the 3D backbone frozen and use a relative-pose head as a lightweight front-end for aggregation and graph refinement.

\paragraph{Feed-forward 3D Geometry Models.}
Feed-forward 3D models replace matching, registration, and optimization with direct geometric prediction.
DUSt3R/MASt3R~\cite{wang2024dust3r,leroy2024mast3r} made pointmaps a general target, and the paradigm has expanded to multi-view, global, permutation-equivariant, dynamic, and prior- or metric-aware reconstruction~\cite{yang2025fast3r,cabon2025must3r,zhang2025flare,wang2025vggt,wang2026pi3,jang2025pow3r,keetha2025mapanything,zhang2025monst3r,chen2025easi3r,chen2026human3r}.
DA3~\cite{lin2025da3} is the most relevant prior for us: built on DINOv2~\cite{oquab2024dinov2}, it predicts spatially consistent geometry through a unified depth-ray target.
Instead of training another foundation model, we ask whether DA3 features can support a control layer for view trust, aggregation, and pair rejection through a small relative-pose head with task-aligned confidence.

\paragraph{Streaming 3D Reconstruction.}
Feed-forward streaming methods either introduce learned state or explicit memory~\cite{wang2025spann3r,wang2025cut3r,wu2025point3r}, adapt transformer aggregation through long-context inference, caches, windows, or token-budget management~\cite{chen2025long3r,lan2025stream3r,zhuo2026streamvggt,li2026wint3r,cheng2026longstream,shen2025fastvggt,yuan2026infinitevggt,lu2026ovggt}, or use test-time training as scene memory~\cite{chen2025ttt3r,elflein2026vggt3,jin2026zipmap,zhang2026loger,xie2026scal3r}.
Concurrent with our work, LingBot-Map~\cite{chen2026lingbotmap} achieves very strong results with a geometric context transformer combining anchor context, pose-reference windows, and trajectory memory, but is trained on a hundred-GPU cluster with large scale internal datasets, so a fair comparison is unpractical.
\modelname is architecturally orthogonal: it introduces no recurrent state, TTT modules, or additional transformers. Instead, it reformulates absolute pose prediction as a system of confidence-weighted relative constraints that aggregate into consistent global trajectories for both streaming and offline inference.

\paragraph{Confidence in 3D Reconstruction.}
Per-pixel or per-correspondence confidences are standard in pointmap and SLAM models~\cite{wang2024dust3r,leroy2024mast3r,wang2025vggt,lin2025da3,teed2021droidslam,teed2023dpvo}, but tied to local uses like depth weighting, correspondence filtering, or optimization residuals.
Our contribution is the \emph{role} of confidence: a pairwise pose confidence decoupled into rotation and translation that serves as loss weight, pairwise-aggregation weight, keyframe-bank utility, and outlier gate in streaming, and as per-pair reliability when fusing pairs in full-context refinement.
This cross-module reuse lets a thin head on the backbone act as a learned front-end for online reconstruction and offline refinement, without hand-designed information matrices or a new memory architecture.

\section{Method}

\begin{figure}[t]
    \vspace{-20pt}
  \centering
  \includegraphics[width=\linewidth]{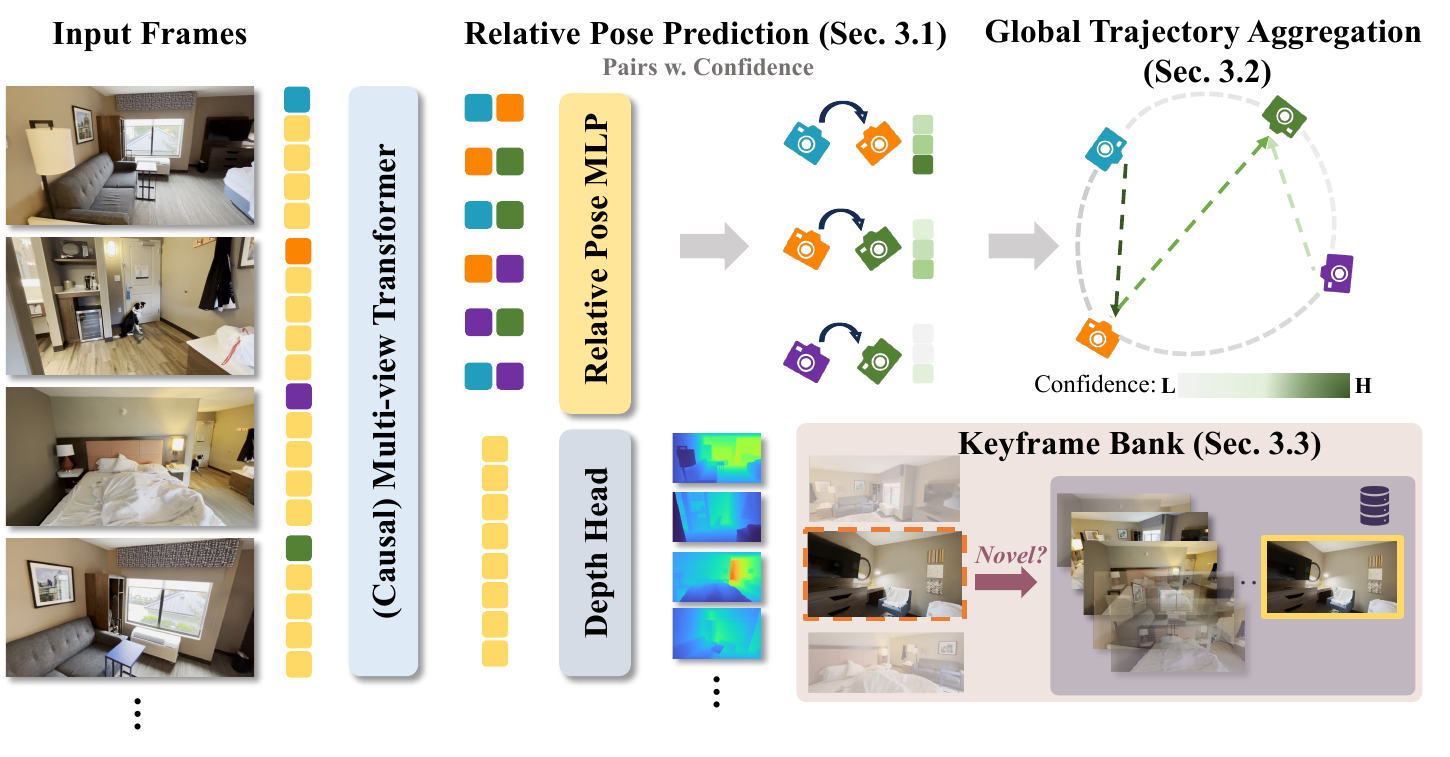}
  \vspace{-20pt}
  \caption{\textbf{Overview of \modelname.} A causal geometry backbone extracts a single camera token from each frame. A lightweight pairwise pose head then predicts directed relative-pose edges from token pairs, along with separate rotation and translation confidences. These confidence-weighted edges are fused into a coherent trajectory, enabling streaming inference with a bounded active keyframe bank.}
  \label{fig:pipeline}
  \vspace{-12pt}
\end{figure}

\definecolor{RedColor}{RGB}{211, 47, 47}    %
\definecolor{GreenColor}{RGB}{56, 142, 60}  %
\definecolor{OrangeColor}{RGB}{245, 124, 0} %

\newcommand{\xmark}{\textcolor{RedColor}{\ding{55}}\xspace}
\newcommand{\cmark}{\textcolor{GreenColor}{\ding{51}}\xspace}
\newcommand{\wcmark}{\textcolor{OrangeColor}{\ding{51}}\xspace}

\begin{wraptable}{r}{0.38\textwidth}
\vspace{-0.5cm}
\centering
\setlength{\tabcolsep}{2.5pt}
\renewcommand{\arraystretch}{0.95}
\resizebox{0.4\textwidth}{!}{%
\begin{tabular}{@{}lccc@{}}
\toprule
\textbf{Attributes} & \textbf{VGGT~\cite{lu2026ovggt}} & \textbf{$\pi^3$~\cite{wang2026pi3}} & \textbf{$R^3$ (Ours)} \\ \midrule
Relative Pose       & \xmark        & \xmark           & \cmark                \\
Relative Loss       & \xmark        & \cmark           & \cmark                \\
All-pair Edges      & \xmark        & \cmark           & \cmark                \\
Learned Confidence  & \xmark        & \xmark           & \cmark                \\ \bottomrule
\end{tabular}
}
\vspace{-0.1cm}
\caption{Pose Paradigm Comparison}
\vspace{-0.3cm}
\label{tab:pose_comparison}
\end{wraptable}

Feed-forward 3D reconstruction from a sequence of $N$ images $\{I_1, \dots, I_N\}$ aims to recover a consistent global 3D representation of the scene. Our model (\modelname) predicts per-frame depth maps $\{D_i\}$ and focal lengths $\{f_i\}$, and derives camera-to-world poses $\{\mathbf{T}_i \in SE(3)\}$ to produce a consistent global point cloud. Unlike prior feed-forward systems that directly regress every $\mathbf{T}_i$ in one global coordinate frame~\cite{wang2025vggt,wang2026pi3,lin2025da3}, \modelname predicts the poses indirectly: it outputs a relative transform between every queried frame pair, and the global trajectory is assembled from these pairwise predictions by a downstream confidence-weighted aggregation. 

As contrasted in \figref{fig:method_comparison} and \tabref{tab:pose_comparison}, \modelname shifts from global anchors to a fully-connected relative pose graph: for each ordered frame pair $(i,j)$ with $i\neq j$, the prediction consists of the relative transform from camera $i$ to camera $j$, together with two scalar confidences --- one for rotation, one for translation. We view these predictions as a \emph{directed pose graph} (\figref{fig:method_comparison}, (c)): frames are nodes, and each queried pair $(i,j)$ becomes a directed edge whose attributes are the predicted relative pose and its two confidences. Since each edge only relates two cameras, it does not require a global coordinate frame. A shared frame is introduced later, when the edges are fused into a trajectory. The overview of our model and inference pipeline is shown in \figref{fig:pipeline}.

\subsection{Relative Pose Prediction with Learned Confidence}\label{sec:rel_pose}

We realize this pairwise framing with a \emph{pairwise pose head}: a single lightweight MLP, shared across all frame pairs, that sits on top of the DA3~\cite{lin2025da3} geometry backbone. The backbone produces one latent camera token $\mathbf{z}_i$ per frame. Given two camera tokens $(\mathbf{z}_i, \mathbf{z}_j)$, the pairwise pose head predicts the relative pose from frame $i$ to frame $j$, together with two scalar confidences:
$$\left(\hat{\mathbf{q}}_{i\rightarrow j},\;\hat{\mathbf{t}}_{i\rightarrow j},\;c^{\mathrm{R}}_{i\rightarrow j},\;c^{\mathrm{T}}_{i\rightarrow j}\right)
=\mathrm{MLP}_{\mathrm{rel}}\!\left([\mathbf{z}_i;\mathbf{z}_j]\right).$$
Here $\hat{\mathbf{q}}_{i\rightarrow j}$ is the relative rotation (unit quaternion), $\hat{\mathbf{t}}_{i\rightarrow j}$ is the translation of camera $j$ expressed in $i$'s coordinate frame, and $c^{\mathrm{R}}_{i\rightarrow j}, c^{\mathrm{T}}_{i\rightarrow j} \!>\! 0$ are the rotation and translation confidences. Each such prediction defines one directed edge $(i, j)$ of the relative-pose graph, with the two confidences as its edge weights. A separate per-frame head predicts the focal length $\hat{f}_i$ from each token $\mathbf{z}_i$.

\paragraph{Relative Versus Absolute Pose Prediction.}
Direct absolute-pose regression ~\cite{wang2025vggt, wang2026pi3, lin2025da3} requires committing to a single global frame. Especially in streaming inference, that frame is fixed from the first several views, and every later pose must be expressed in the same arbitrary coordinate choice. A pairwise edge instead predicts the displacement between its two endpoint cameras. This displacement is not a camera's accumulated displacement from a fixed origin. The confidence-weighted loss below then learns how much each pair should contribute, so weakly constrained long-baseline pairs need not be trusted as much as reliable ones. The pairwise head also exposes $O(N^2)$ supervised pose constraints from an $N$-frame training clip while reusing one shared MLP across all queried pairs.

\paragraph{Confidence Design.}
Rotation and translation exhibit different failure modes: translation is often less reliable when geometric cues are weak, whereas rotation can remain well constrained. A single shared confidence cannot express this asymmetry, so the head predicts separate reliability estimates for rotation and translation. The next sections describe how these two confidences enter the training loss and the trajectory aggregation. \appref{app:pose_supervision} gives a more detailed comparison of pose-supervision choices and analyzes the learned confidences as pair-reliability estimates.

\subsection{Global Trajectory Aggregation}\label{sec:abs_pose}

The trained pairwise pose head produces local edges, but downstream tasks such as depth fusion and point-cloud construction need every camera expressed in a single shared frame. Aggregation closes this gap by stitching the edges into a global trajectory.

We first describe aggregation in the causal streaming setting. Frame $1$ defines the origin and orientation of the shared coordinate system, i.e., $\mathbf{T}_1 = I$. Frame $2$ can then be placed directly from pair $(1,2)$. For a later frame $j$, each available reference frame $i$ proposes one candidate pose for $j$: we take the already-estimated pose of frame $i$ and compose it with the relative transform predicted for pair $(i,j)$,
$$\mathbf{q}_{j}^{(i)} = \mathbf{q}_i \otimes \hat{\mathbf{q}}_{i\rightarrow j}, \quad \mathbf{t}_{j}^{(i)} = \mathbf{t}_i + \mathbf{q}_i(\hat{\mathbf{t}}_{i\rightarrow j}),$$
where $\mathbf{q}_i(\cdot)$ rotates a vector by $\mathbf{q}_i$. We call the available references for frame $j$ the reference set $\mathcal{R}_j$. In streaming, $\mathcal{R}_j$ is the active context $\mathcal{C}_t$ defined below: it is not the full history, but the subset of earlier frames currently kept in memory.

The final pose is obtained by confidence-weighted fusion over these candidate poses. Rotation and translation use separate softmax weights from $c^{\mathrm{R}}_{i\rightarrow j}$ and $c^{\mathrm{T}}_{i\rightarrow j}$, so unreliable pair predictions contribute less to the corresponding component. When computation permits, we fuse all references. For efficiency, we may instead restrict fusion to the top-$K$ references ranked by averaged confidence $\bar{c}_{i\rightarrow j} = \frac{1}{2}(c^{\mathrm{R}}_{i\rightarrow j} + c^{\mathrm{T}}_{i\rightarrow j})$; using all references is recovered when $K \ge |\mathcal{R}_j|$. With $\mathcal{N}_K(j)$ denoting the retained references, the fused pose is
$$\mathbf{t}_j = \sum_{i\in\mathcal{N}_K(j)} \tilde{c}^{\mathrm{T}}_{i\rightarrow j} \mathbf{t}_{j}^{(i)}, \quad \mathbf{q}_j = \mathrm{Norm}\!\left(\sum_{i\in\mathcal{N}_K(j)} \tilde{c}^{\mathrm{R}}_{i\rightarrow j} \mathbf{q}_{j}^{(i)}\right),$$
where $\tilde{c}^{\mathrm{R}}_{i\rightarrow j}, \tilde{c}^{\mathrm{T}}_{i\rightarrow j}$ are the softmax normalizations of $c^{\mathrm{R}}_{i\rightarrow j}, c^{\mathrm{T}}_{i\rightarrow j}$ over $\mathcal{N}_K(j)$, and $\mathrm{Norm}(\cdot)$ renormalizes its argument to a unit quaternion. Candidate quaternions are sign-aligned before averaging.

\paragraph{Confidence-weighted Fusion.}
The aggregation weights come from the predicted rotation and translation confidences, not from a fixed averaging rule. The top-$K$ neighborhood is only an efficiency cap when many references are available; the all-reference version is evaluated in \appref{app:abl_aggregation}.

\subsection{Causal Streaming and Full-Context Inference}\label{sec:inference}
\subsubsection{Streaming Inference with an Active Context}\label{sec:online}

In streaming mode, frames arrive sequentially; each incoming frame is paired only with a small active context rather than the full history. The active context contains frame $1$, which fixes the trajectory origin, and a dynamic keyframe bank of previously accepted frames: $\mathcal{C}_t = \{1\} \cup \mathcal{B}_t$. For incoming frame $j$, the pairwise pose head is evaluated only on edges between $j$ and frames in $\mathcal{C}_t$, and aggregation uses $\mathcal{R}_j = \mathcal{C}_t$. This works because frames enter the bank only after their poses have been estimated.
The bank $\mathcal{B}_t$ is managed by two rules: adding feature-novel frames and removing the least useful keyframes when the bank is full.

\paragraph{Adding Keyframes.}
Frame $j$ is admitted to $\mathcal{B}_t$ only if its pre-attention backbone token is sufficiently different from the current bank:
$$\max_{i\in\mathcal{B}_t}\cos(\mathbf{tok}_i,\mathbf{tok}_j)<\tau,$$
where $\mathbf{tok}_j$ is the average of $j$'s backbone encoder tokens before any cross-frame interaction, and $\tau$ is the novelty threshold. Since the backbone is pretrained for 3D reconstruction, this token provides a cue for local geometric and appearance redundancy.

\paragraph{Culling.}
When $\mathcal{B}_t$ reaches its capacity $M_{\max}$, the entry with the lowest utility $u_j = d_j \, c_j$ is evicted. Here $d_j = \min_{i\in\mathcal{B}_t\setminus\{j\}}\!\big(1-\cos(\mathbf{tok}_i,\mathbf{tok}_j)\big)$ is the token-level distinctiveness of frame $j$ from the closest other bank entry, and $c_j$ is the strongest pair confidence between $j$ and the rest of the bank (\appref{app:online_details}). The first frame is excluded from eviction.

The bank mirrors classical keyframe sparsification~\cite{klein2007ptam,murartal2015orbslam,murartal2017orbslam2,engel2017dso,campos2021orbslam3}, but replaces hand-designed motion and covisibility rules with token novelty and pose-head confidence.

\subsubsection{Full-context Inference}
When the complete clip is available, we can use the same checkpoint in a full-context mode: removing the causal attention mask at test time lets the backbone see all frames at once, so the pairwise pose head can be queried on every pair $(i, j)$ with $i \neq j$. We initialize the trajectory with a causal streaming pass and then run one lightweight confidence-weighted pose-graph refinement over the predicted relative-pose edges. This refinement is inexpensive because its edges carry only relative-pose residuals and scalar confidences; it does not run bundle adjustment, point reprojection, or depth optimization. Additional implementation details for the keyframe bank and refinement solver are given in \appref{app:online_details}.

\subsection{Training Objective}\label{sec:training}
The pairwise pose head and depth head are supervised with confidence-weighted residual losses. The total objective is $\mathcal{L} = \mathcal{L}_{\mathrm{cam}} + \mathcal{L}_{\mathrm{depth}}$.

\paragraph{Confidence-aware Camera Loss.}
For each pair $(i, j)$ with $i \neq j$, rotation and translation are supervised by confidence-weighted $L_1$ residuals:
$$\mathcal{L}_{\mathrm{rot}}(i,j)=c^{\mathrm{R}}_{i\rightarrow j}\,\ell^{L_1}_{\mathrm{rot}}(i,j)-\alpha\log c^{\mathrm{R}}_{i\rightarrow j}, \quad \mathcal{L}_{\mathrm{trans}}(i,j)=c^{\mathrm{T}}_{i\rightarrow j}\,\ell^{L_1}_{\mathrm{trans}}(i,j)-\alpha\log c^{\mathrm{T}}_{i\rightarrow j},$$
where $\ell^{L_1}_{\mathrm{rot}}, \ell^{L_1}_{\mathrm{trans}}$ are $L_1$ residuals against the ground-truth rotation and translation (\appref{app:pose_supervision}). In the causal setting, $\mathcal{L}_{\mathrm{cam}}$ averages these terms over past-to-current ordered pairs and adds a plain per-frame $L_1$ term for the focal-length head. The two parts of each loss work in tension: a large residual makes the product $c\,\ell$ costly; consequently, the optimizer shrinks $c$ on inaccurate pairs, while the $-\log c$ regularizer ($\alpha=0.2$) keeps $c$ from collapsing to zero. As training proceeds, the model assigns higher $c^{\mathrm{R}}, c^{\mathrm{T}}$ to pairs with small residuals and lower confidence to pairs with large residuals.

\paragraph{Depth Loss.}
The depth head is supervised in median-normalized space: each prediction $D$ and target $D^*$ is divided by its own per-frame median to remove scene-scale ambiguity, yielding $\tilde D$ and $\tilde D^*$. The supervision target $D^*$ depends on the data source: ground-truth depth on synthetic scenes, and the depth output of a frozen pretrained DA3~\cite{lin2025da3} model on real-world sequences. The loss is
$$\mathcal{L}_{\mathrm{depth}}= \frac{1}{|\mathcal{M}|}\sum_{p\in\mathcal{M}} \Big(\Sigma_p\,\big|\tilde D(p)-\tilde D^*(p)\big|-\alpha\log\Sigma_p\Big),$$
where $\Sigma_p$ is the learned per-pixel depth confidence and the supervision mask $\mathcal{M}$ uses ground-truth validity for synthetic data and DA3-teacher confidence for real-world data.

\paragraph{Confidence-weighted Supervision.}
A fixed-weight camera loss $\lambda_{\mathrm{R}}\ell_{\mathrm{rot}}+\lambda_{\mathrm{T}}\ell_{\mathrm{trans}}$ uses the same $\lambda_{\mathrm{R}}, \lambda_{\mathrm{T}}$ for every pair, even when the right balance varies across pairs. Our confidence-weighted loss adapts this balance during training, upweighting confident pairs and downweighting unreliable ones; the depth loss applies the same idea per pixel.

Architecture updates, optimization schedules, datasets, and other training details are given in \appref{app:training_details}.

\newcommand{\placeholder}[1]{\textcolor{red}{[#1]}}

\section{Experiments}
We evaluate \modelname across four dimensions: (i) camera pose accuracy, (ii) dense point-map reconstruction quality, (iii) scalability for long sequences, and (iv) robustness to distractor frames.

Unless otherwise specified, all evaluations follow the standard protocols of the respective benchmarks. Our single causal checkpoint runs in the default streaming mode with bounded memory; when the complete clip is available, we also report a lightweight full-context switch obtained by removing the causal mask at test time and running one confidence-weighted pose-graph refinement. To assess scalability, all methods are tested under a fixed 48\,GiB GPU memory budget, where ``OOM'' indicates failure to meet this constraint.

\subsection{Camera Pose Estimation}
\label{sec:pose}

We evaluate camera pose accuracy on Sintel~\cite{butler2012sintel}, TUM-dynamics~\cite{sturm2012tum}, and ScanNet~\cite{dai2017scannet}, comparing streaming \modelname against existing streaming baselines and also reporting full-sequence context results.

\begin{table}[!ht]
\vspace{-12pt}
\centering
\captionsetup{aboveskip=2pt,belowskip=2pt}
\caption{\textbf{Camera pose estimation on offline (Top) and online (Bottom) settings.} ATE / RPE-T / RPE-R, all lower is better; RPE-R is in degrees. VGGT, DA3-Large, and \modelname (full context) are full-sequence references. \textbf{Bold}/underlined marks best/second-best within each block. $^{\dagger}$ marks offline methods. $^{\ast}$ indicates values taken from the paper.
}
\label{tab:pose_online}
\footnotesize
\setlength{\tabcolsep}{2.4pt}
\renewcommand{\arraystretch}{0.85}
\resizebox{0.88\linewidth}{!}{%
\begin{tabular}{l c | ccc | ccc | ccc}
\toprule
& & \multicolumn{3}{c|}{Sintel (50 frames)} & \multicolumn{3}{c|}{TUM-dynamics (90 frames)} & \multicolumn{3}{c}{ScanNet (90 frames)} \\
Method & \#Params & ATE$\downarrow$ & RPE-T$\downarrow$ & RPE-R$\downarrow$ & ATE$\downarrow$ & RPE-T$\downarrow$ & RPE-R$\downarrow$ & ATE$\downarrow$ & RPE-T$\downarrow$ & RPE-R$\downarrow$ \\
\midrule
VGGT\offmark~\cite{wang2025vggt}             & 1.26B  & \third{0.172}  & \third{0.061}  & \second{0.471} & \best{0.012}   & \second{0.010} & \best{0.309}   & \best{0.035}   & \best{0.015}   & \best{0.376}   \\
DA3-Large\offmark~\cite{lin2025da3}          & 385M   & \second{0.140} & \second{0.059} & \best{0.450}   & \second{0.013} & \third{0.011}  & \second{0.310} & \third{0.039}  & \second{0.017} & \third{0.590}  \\
\modelnamebf\offmark          & \best{372M}   & \best{0.130}   & \best{0.047}   & \third{0.523}  & \best{0.012}   & \best{0.009}   & \third{0.318}  & \second{0.037} & \best{0.015}   & \second{0.472} \\
\midrule
Spann3R~\cite{wang2025spann3r}                & ---    & 0.329 & 0.110 & 4.471 & 0.056 & 0.021 & 0.591 & 0.096 & 0.023 & 0.661 \\
CUT3R~\cite{wang2025cut3r}                    & 793M   & 0.213 & \third{0.066} & 0.621 & 0.046 & 0.015 & 0.473 & 0.099 & 0.022 & \third{0.600} \\
Point3R~\cite{wu2025point3r}                  & ---    & 0.351 & 0.128 & 1.822 & 0.075 & 0.029 & 0.642 & 0.106 & 0.035 & 1.946 \\
StreamVGGT~\cite{zhuo2026streamvggt}          & 1.26B  & 0.251 & 0.149 & 1.894 & 0.061 & 0.033 & 3.209 & 0.161 & 0.057 & 3.647 \\
STream3R~\cite{lan2025stream3r}               & 1.26B  & 0.213 & 0.076 & 0.868 & \second{0.026} & \third{0.013} & \second{0.330} & \second{0.052} & \second{0.021} & 0.850 \\
TTT3R~\cite{chen2025ttt3r}                    & 793M   & 0.201 & \best{0.063} & \second{0.617} & \third{0.028} & \second{0.012} & \third{0.379} & 0.064 & \second{0.021} & \second{0.592} \\
ZipMap-stream$^{\ast}$~\cite{jin2026zipmap}   & 1.40B  & \second{0.159} & \second{0.065} & 0.750 & --- & --- & --- & --- & --- & --- \\
\modelnamebf                                  & \best{372M}   & \best{0.115} & 0.068 & \best{0.548} & \best{0.018} & \best{0.010} & \best{0.320} & \best{0.038} & \best{0.016} & \best{0.480} \\
\bottomrule
\end{tabular}%
}
\vspace{-0.8cm}
\end{table}

As shown in~\tabref{tab:pose_online}, streaming \modelname provides the strongest overall pose accuracy among streaming methods while using a compact $372$M-parameter model. This is substantially smaller than most recent streaming reconstruction baselines with reported model sizes, yet it improves global trajectory accuracy across all three benchmarks, with only one short-term translation metric slightly trailing the best baseline. When the complete sequence is available, the same checkpoint can be switched to full-context inference and remains competitive with DA3/VGGT references without retraining. The slightly worse ATE on Sintel is due to the outlier sequence \texttt{sintel-temple-3}.

\vspace{-0.2cm}
\subsection{Point-Map Reconstruction}
\label{sec:pointmap_sparse}
Following CUT3R~\cite{wang2025cut3r}, we evaluate dense reconstruction on 7-Scenes~\cite{shotton2013scenecoord} and NRGBD~\cite{azinovic2022neuralrgbd} with uniform keyframe sampling (strides 200 and 100). In \tabref{tab:pointmap_sparse}, streaming \modelname leads on 7-Scenes, showing that the pose estimates translate into accurate geometry under sparse sampling, and remains competitive with the strongest streaming baseline on NRGBD. The full-context switch improves the primary mean Acc/Comp values on both datasets from the same checkpoint.

\begin{table}[!ht]
\vspace{-12pt}
\centering
\captionsetup{aboveskip=2pt,belowskip=2pt}
\caption{\textbf{Sparse-view point-map reconstruction.} Acc and Comp are lower-is-better; NC is higher-is-better. VGGT, DA3-Large, and \modelname (full context) are full-sequence references. \textbf{Bold}/underlined marks best/second-best within each block. $^{\dagger}$ marks offline methods. 
}
\label{tab:pointmap_sparse}
\footnotesize
\setlength{\tabcolsep}{2.6pt}
\renewcommand{\arraystretch}{0.85}
\resizebox{0.86\linewidth}{!}{%
\begin{tabular}{l cccccc cccccc}
\toprule
& \multicolumn{6}{c}{7-Scenes~\cite{shotton2013scenecoord}} & \multicolumn{6}{c}{NRGBD~\cite{azinovic2022neuralrgbd}} \\
\cmidrule(lr){2-7} \cmidrule(lr){8-13}
& \multicolumn{2}{c}{Acc$\downarrow$} & \multicolumn{2}{c}{Comp$\downarrow$} & \multicolumn{2}{c}{NC$\uparrow$} & \multicolumn{2}{c}{Acc$\downarrow$} & \multicolumn{2}{c}{Comp$\downarrow$} & \multicolumn{2}{c}{NC$\uparrow$} \\
\cmidrule(lr){2-3} \cmidrule(lr){4-5} \cmidrule(lr){6-7} \cmidrule(lr){8-9} \cmidrule(lr){10-11} \cmidrule(lr){12-13}
Method & Mean & Med. & Mean & Med. & Mean & Med. & Mean & Med. & Mean & Med. & Mean & Med. \\
\midrule
VGGT\offmark~\cite{wang2025vggt}             & \third{0.087}  & \second{0.039} & \third{0.091}  & \third{0.039}  & \best{0.787}   & \best{0.890}   & \third{0.073}  & \best{0.018}   & \third{0.077}  & \second{0.021} & \best{0.910}   & \best{0.990}   \\
DA3-Large\offmark~\cite{lin2025da3}          & \second{0.086} & \best{0.034}   & \second{0.088} & \second{0.033} & \third{0.756}  & \third{0.873}  & \second{0.045} & \second{0.020} & \second{0.061} & \third{0.025}  & \second{0.896} & \second{0.988} \\
\modelnamebf\offmark (full context)          & \best{0.081}   & \third{0.043}  & \best{0.070}   & \best{0.027}   & \second{0.770} & \second{0.881} & \best{0.038}   & \second{0.020} & \best{0.044}   & \best{0.019}   & \third{0.887}  & \best{0.990}   \\
\midrule
Spann3R~\cite{wang2025spann3r}       & 0.298          & 0.226          & 0.205          & 0.112          & 0.650          & 0.730          & 0.416          & 0.323          & 0.417          & 0.285          & 0.684          & 0.789          \\
CUT3R~\cite{wang2025cut3r}           & \third{0.126}  & \third{0.047}  & 0.154          & \best{0.031}   & 0.727          & 0.834          & 0.099          & \third{0.031}  & 0.076          & \third{0.026}  & 0.837          & 0.971          \\
STream3R~\cite{lan2025stream3r}      & \second{0.122} & \second{0.044} & \second{0.110} & \second{0.038} & \third{0.746}  & \third{0.856}  & \second{0.057} & \best{0.014}   & \best{0.028}   & \best{0.013}   & \best{0.910}   & \best{0.993}   \\
StreamVGGT~\cite{zhuo2026streamvggt} & 0.129          & 0.056          & \third{0.115}  & 0.041          & \second{0.751} & \second{0.865} & \third{0.084}  & 0.044          & \third{0.074}  & 0.041          & \third{0.861}  & \third{0.986}  \\
\modelnamebf                         & \best{0.092}   & \best{0.038}   & \best{0.093}   & \third{0.040}  & \best{0.755}   & \best{0.868}   & \best{0.047}   & \second{0.022} & \second{0.050} & \second{0.020} & \second{0.883} & \second{0.991} \\
\bottomrule
\end{tabular}}
\vspace{-1cm}
\end{table}

\subsection{Long-Sequence Scalability}
\label{sec:longseq}
\begin{figure}[H]
    \vspace{-1.5em}
  \centering
  \begin{subfigure}[b]{0.44\linewidth}
    \centering
    \includegraphics[width=\linewidth]{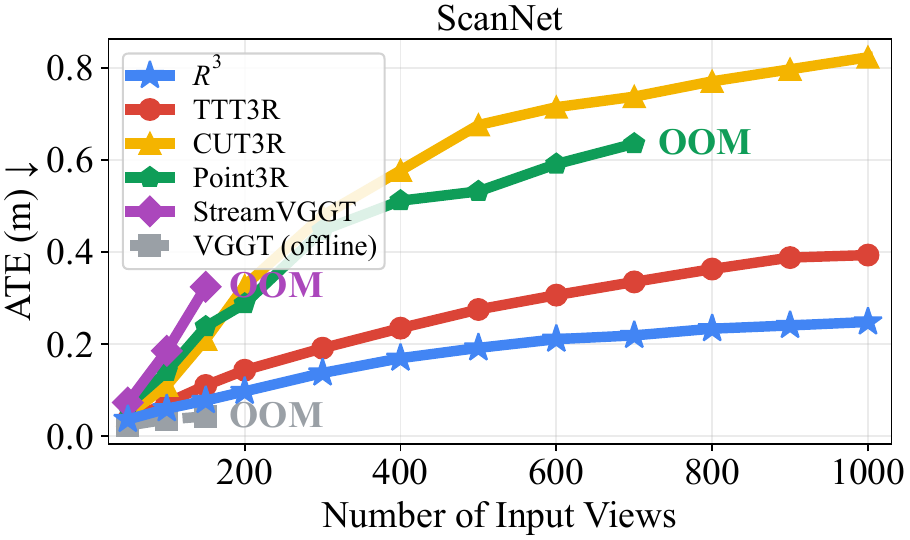}
    \vspace{-14pt}
  \end{subfigure}\hspace{6pt}%
  \begin{subfigure}[b]{0.44\linewidth}
    \centering
    \includegraphics[width=\linewidth]{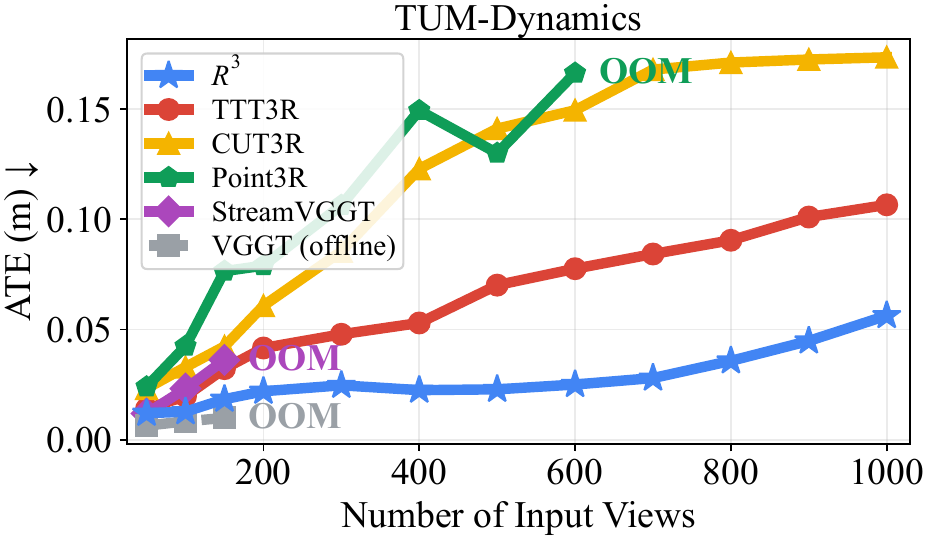}
    \vspace{-14pt}
  \end{subfigure}%
  \vspace{-4pt}
  \caption{\textbf{Pose accuracy scaling on long sequences.} We plot ATE for ScanNet~\cite{dai2017scannet} and TUM-dynamics~\cite{sturm2012tum} as the number of input frames increases. While several streaming baselines exhibit cumulative drift or trigger out-of-memory (OOM) failures, \modelname maintains stable trajectory estimation.}
  \label{fig:ate_pose_long}
   \vspace{-10pt}
\end{figure}

\begin{figure}[!t]
\vspace{-20pt}
  \centering
  \captionsetup{aboveskip=2pt,belowskip=0pt}
  \includegraphics[width=1\linewidth]{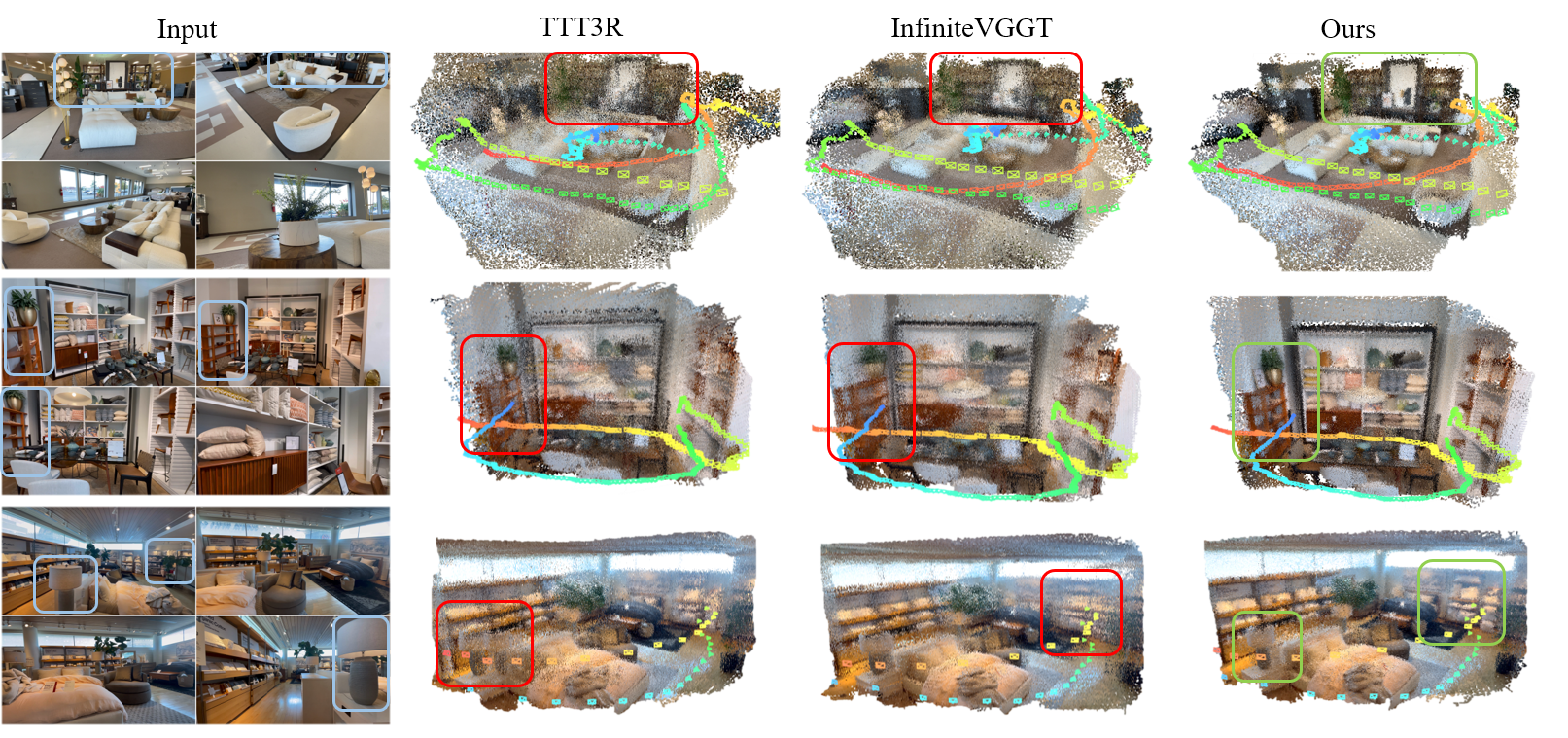}
  \caption{\textbf{Long streaming comparison.} Qualitative in-the-wild results show that \modelname maintains more consistent trajectories and point-map alignments over hundreds of frames than baselines.}
  \label{fig:long_seq_comparison}
\vspace{-12pt}
\end{figure}

We evaluate long-sequence behavior on dense reconstruction of 7-Scenes at sequence lengths up to 1000 frames (\tabref{tab:scalability}) and on ScanNet / TUM-dynamics for pose accuracy (\figref{fig:ate_pose_long}). \modelname processes each stream in a \emph{single pass without reset}. On 7-Scenes, streaming baselines lose Acc/Comp at 1000 frames, while \modelname remains roughly flat. On ScanNet / TUM-dynamics, its ATE grows gradually as views are added but avoids the stronger drift observed in the baselines. We attribute this stability to confidence-weighted aggregation over the keyframe bank. \figref{fig:long_seq_comparison} shows the same trend qualitatively on in-the-wild clips. FPS and GPU-memory curves under the same protocol are deferred to \appref{app:compute_cost}.

\begin{table}[H]
\centering
\vspace{-10pt}
\caption{\textbf{Long-sequence reconstruction on 7-Scenes.} Mean values at sequence lengths $200/500/1000$. Acc / Comp are lower-is-better; NC is higher-is-better. ``OOM'' marks methods that exceed the 48\,GiB GPU memory budget. \textbf{Bold}/underlined marks best/second-best per metric column.}
\label{tab:scalability}
\small
\setlength{\tabcolsep}{3pt}
\renewcommand{\arraystretch}{0.85}
\resizebox{0.78\linewidth}{!}{%
\begin{tabular}{l ccc ccc ccc}
\toprule
& \multicolumn{3}{c}{Length 200} & \multicolumn{3}{c}{Length 500} & \multicolumn{3}{c}{Length 1000} \\
\cmidrule(lr){2-4} \cmidrule(lr){5-7} \cmidrule(lr){8-10}
Method & Acc$\downarrow$ & Comp$\downarrow$ & NC$\uparrow$ & Acc$\downarrow$ & Comp$\downarrow$ & NC$\uparrow$ & Acc$\downarrow$ & Comp$\downarrow$ & NC$\uparrow$ \\
\midrule
Spann3R~\cite{wang2025spann3r}              & 0.215          & 0.122          & 0.535          & 0.343          & 0.154          & 0.515          & 0.340          & 0.154          & 0.508 \\
CUT3R~\cite{wang2025cut3r}                  & 0.087          & 0.045          & 0.566          & 0.194          & 0.092          & 0.527          & 0.240          & 0.102          & 0.513 \\
Point3R~\cite{wu2025point3r}                & 0.041          & \second{0.023} & 0.579          & \third{0.056}  & 0.031          & \third{0.555}  & \third{0.068}  & \second{0.025} & \third{0.533} \\
TTT3R~\cite{chen2025ttt3r}                  & \second{0.027} & \second{0.023} & \third{0.582}  & 0.065          & \third{0.030}  & 0.552          & 0.126          & 0.050          & 0.525 \\
StreamVGGT~\cite{zhuo2026streamvggt}        & \third{0.038}  & \third{0.029}  & \second{0.583} & OOM            & OOM            & OOM            & OOM            & OOM            & OOM \\
InfiniteVGGT~\cite{yuan2026infinitevggt}    & 0.046          & 0.031          & \third{0.582}  & \second{0.040} & \second{0.024} & \second{0.561} & \second{0.061} & \third{0.035}  & \second{0.537} \\
\midrule
\modelnamebf                                & \best{0.021}   & \best{0.018}   & \best{0.600}   & \best{0.022}   & \best{0.017}   & \best{0.562}   & \best{0.022}   & \best{0.017}   & \best{0.538} \\
\bottomrule
\end{tabular}%
}
\end{table}

\begin{table}[!ht]
\vspace{-0.8cm}
\centering
\caption{\textbf{Long-sequence camera trajectory accuracy on DL3DV-Benchmark.} ATE-norm / ATE-RMSE / Rot RMSE, all lower is better; ATE-norm is in \%, Rot RMSE is in degrees. Win rate counts scenes (out of $25$) where each method attains the lowest ATE-norm. \textbf{Bold} marks the best per column.}
\label{tab:dl3dv_pose}
\small
\setlength{\tabcolsep}{6pt}
\begin{tabular}{l c c c c}
\toprule
Method & ATE-norm (\%)$\downarrow$ & ATE-RMSE$\downarrow$ & Rot RMSE ($^\circ$)$\downarrow$ & Win rate \\
\midrule
TTT3R~\cite{chen2025ttt3r}                & 4.91 & 0.601 & 5.65 & 1/25 \\
InfiniteVGGT~\cite{yuan2026infinitevggt}  & 2.85 & 0.367 & 3.06 & 0/25 \\
\midrule
\modelnamebf                              & \textbf{1.16} & \textbf{0.155} & \textbf{1.79} & \textbf{24/25} \\
\bottomrule
\end{tabular}
\end{table}

We further test pose-only trajectory accuracy on a subset of DL3DV-Benchmark~\cite{ling2024dl3dv} ($304$--$439$ frames), which contains wider camera baselines and outdoor scenes beyond the mostly indoor TUM-dynamics/ScanNet setting. TTT3R uses a reset interval because it outperforms no-reset inference; the full protocol is provided in \appref{app:dl3dv_pose}. As shown in \tabref{tab:dl3dv_pose}, \modelname wins ATE-norm on $24/25$ scenes and sharply lowers mean ATE-norm versus the strongest streaming baseline.

\newpage
\subsection{Robustness via Confidence Gating}
\label{sec:robustness}

\begin{wraptable}[11]{r}{0.45\linewidth}
\vspace{-0.6cm}
\centering
\caption{\textbf{Robustness with distractors.} Mean over Small/Medium/Large distractor settings ($10$ seeds each); SR denotes rejection success. %
}
\label{tab:robustness}
\footnotesize
\setlength{\tabcolsep}{3pt}
\renewcommand{\arraystretch}{0.85}
\begin{tabular}{@{}lcccc@{}}
\toprule
\multirow{2}{*}{Method} & \multicolumn{2}{c}{ETH3D~\cite{schops2017eth3d}} & \multicolumn{2}{c}{RobustNeRF~\cite{sabour2023robustnerf}} \\
\cmidrule(lr){2-3} \cmidrule(l){4-5}
& ATE$\downarrow$ & SR$\uparrow$ & ATE$\downarrow$ & SR$\uparrow$ \\
\midrule
RobustVGGT-A\offmark        & 0.733 & 0.914 & 0.138 & 0.641 \\
RobustVGGT-F\offmark        & 0.763 & 0.985 & \best{0.138} & 0.586 \\
\modelname               & 0.998 & --    & 0.199 & --    \\
\modelnamebf + reject     & \best{0.244} & \best{1.000} & 0.152 & \best{0.986} \\
\bottomrule
\vspace{-3.0cm}
\end{tabular}
\vspace{-10.5cm}
\end{wraptable}

The learned confidences is able to further function as an effective outlier gate. Specifically, when the mean confidence of a new frame $j$ against the active context $\mathcal{C}_t$ falls below a calibrated baseline, we invalidate its KV-cache entries, skip bank admission, and suppress its pose estimation. This mechanism enables \modelname to handle transient failures such as motion blur, occlusions, or sudden scene cuts without polluting the keyframe bank. We provide details on calibration, thresholds, and our segment-reset strategy in \appref{app:online_details}.

Following the Robust-VGGT~\cite{han2025emergent} protocol, we interleave distractor frames into the input stream while maintaining temporal order. As shown in \tabref{tab:robustness}, \modelname identifies distractors in a single online pass, whereas Robust-VGGT requires a multi-pass offline detection stage. This validates that our reliability signal effectively doubles as an out-of-scene detector, enabling robust streaming reconstruction using a unified rule across experiments.

\section{Ablation Study}\label{sec:ablation}

\subsection{Effectiveness of Relative-Pose Formulation}
\label{sec:relpose_vs_direct}

To isolate the intrinsic benefits of the relative-pose formulation, we conduct a controlled comparison against a direct global-pose regression baseline. Within each setting, we keep the backbone architecture, training data, and optimization budget identical, varying only the prediction target and its associated loss. The streaming block uses an absolute-pose-pretrained backbone, which biases the setup toward the direct baseline rather than our relative-pose head.

\begin{table}[h]
\vspace{-10pt}
\centering
\caption{\textbf{Relative-pose prediction objective.} ATE / RPE-T / RPE-R, all lower is better. Within each block, the controlled variants share the same backbone setting and differ in the pose prediction target and loss. \textbf{Bold} marks the best result within each block. $^{\dagger}$ marks full-sequence methods.}
\label{tab:pi3_protocol}
\small
\setlength{\tabcolsep}{4pt}
\renewcommand{\arraystretch}{0.85}
\begin{tabular}{l ccc ccc}
\toprule
& \multicolumn{3}{c}{Sintel~\cite{butler2012sintel}} & \multicolumn{3}{c}{ScanNet~\cite{dai2017scannet}} \\
\cmidrule(lr){2-4} \cmidrule(lr){5-7}
Variant & ATE$\downarrow$ & RPE-T$\downarrow$ & RPE-R$\downarrow$ & ATE$\downarrow$ & RPE-T$\downarrow$ & RPE-R$\downarrow$ \\
\midrule
\multicolumn{7}{l}{\textit{Full-context (VGGT init, $64$k iters)}} \\
Backbone direct (Pi3 loss)\offmark   & 0.1583 & 0.0841 & 0.9492 & 0.0683 & 0.0195 & 0.5143 \\
Backbone + our rel. target/loss\offmark       & \best{0.1370} & \best{0.0757} & \best{0.7282} & \best{0.0604} & \best{0.0189} & \best{0.5039} \\
\midrule
\multicolumn{7}{l}{\textit{Streaming ($5$k iters, abs-pose-pretrained backbone, pose heads reset)}} \\
Backbone direct (VGGT abs loss) & \best{0.141} & 0.079 & 1.56 & 0.063 & 0.026 & 1.33 \\
Backbone + our rel. target/loss          & 0.146 & \best{0.066} & \best{0.67} & \best{0.052} & \best{0.021} & \best{0.93} \\
\bottomrule
\end{tabular}
\vspace{-10pt}
\end{table}

As shown in~\tabref{tab:pi3_protocol}, the relative formulation improves most metrics in both full-context and streaming scenarios, especially RPE and ScanNet ATE; the direct baseline remains slightly better on Sintel streaming ATE. Overall, the results indicate that supervised relative-pose regression is a strong geometric prior for long-range sequences, even before any complex aggregation is applied.
Additional ablations on full-attention fine-tuning, aggregation strategy, and keyframe admission are in \appref{app:additional_ablations}.

\section{Conclusion}

We presented \modelname, a feed-forward 3D reconstruction framework that reformulates global-frame regression as pairwise relative-pose regression. Our approach employs a lightweight MLP to predict relative motion alongside decoupled confidences for rotation and translation. This confidence signal acts as a unified primitive: it weights training losses, guides pose aggregation at inference, and drives keyframe-bank management. Experimental results demonstrate that \modelname is competitive with the baselines, even on sequences exceeding thousands of frames. By offloading the coordinate-frame choice to a relative pose aggregation step, \modelname demonstrates that local geometric relations are a more natural and scalable learning target for feed-forward 3D reconstruction.

\section*{Acknowledgments}

We gratefully thank Siyuan Bian and Weiwei Xu for their insightful discussions, and the members of Jun Gao Lab and Inception3D Lab for their support throughout this project.

\bibliographystyle{plainnat}
\bibliography{references}

@inproceedings{wang2024dust3r,
  title     = {{DUSt3R}: Geometric {3D} Vision Made Easy},
  author    = {Wang, Shuzhe and Leroy, Vincent and Cabon, Yohann and Chidlovskii, Boris and Revaud, Jerome},
  booktitle = {Proceedings of the IEEE/CVF Conference on Computer Vision and Pattern Recognition (CVPR)},
  year      = {2024}
}

@inproceedings{leroy2024mast3r,
  title     = {Grounding Image Matching in {3D} with {MASt3R}},
  author    = {Leroy, Vincent and Cabon, Yohann and Revaud, J{\'e}r{\^o}me},
  booktitle = {European Conference on Computer Vision (ECCV)},
  year      = {2024}
}

@inproceedings{yang2025fast3r,
  title     = {{Fast3R}: Towards {3D} Reconstruction of {1000+} Images in One Forward Pass},
  author    = {Yang, Jianing and Sax, Alexander and Liang, Kevin J. and Henaff, Mikael and Tang, Hao and Cao, Ang and Chai, Joyce and Meier, Franziska and Feiszli, Matt},
  booktitle = {Proceedings of the IEEE/CVF Conference on Computer Vision and Pattern Recognition (CVPR)},
  year      = {2025}
}

@inproceedings{wang2025vggt,
  title     = {{VGGT}: Visual Geometry Grounded Transformer},
  author    = {Wang, Jianyuan and Chen, Minghao and Karaev, Nikita and Vedaldi, Andrea and Rupprecht, Christian and Novotny, David},
  booktitle = {Proceedings of the IEEE/CVF Conference on Computer Vision and Pattern Recognition (CVPR)},
  year      = {2025}
}

@inproceedings{wang2026pi3,
  title     = {{$\pi^3$}: Permutation-Equivariant Visual Geometry Learning},
  author    = {Wang, Yifan and Zhou, Jianjun and Zhu, Haoyi and Chang, Wenzheng and Zhou, Yang and Li, Zizun and Chen, Junyi and Pang, Jiangmiao and Shen, Chunhua and He, Tong},
  booktitle = {International Conference on Learning Representations (ICLR)},
  year      = {2026},
  note      = {arXiv:2507.13347}
}

@article{lin2025da3,
  title   = {Depth Anything 3: Recovering the Visual Space from Any Views},
  author  = {Lin, Haotong and Chen, Sili and Liew, Jun Hao and Chen, Donny Y. and Li, Zhenyu and Shi, Guang and Feng, Jiashi and Kang, Bingyi},
  journal = {arXiv preprint arXiv:2511.10647},
  year    = {2025}
}

@article{keetha2025mapanything,
  title   = {{MapAnything}: Universal Feed-Forward Metric {3D} Reconstruction},
  author  = {Keetha, Nikhil and M{\"u}ller, Norman and Sch{\"o}nberger, Johannes and Porzi, Lorenzo and Zhang, Yuchen and Fischer, Tobias and Knapitsch, Arno and Zauss, Duncan and Weber, Ethan and Antunes, Nelson and Luiten, Jonathon and Lopez-Antequera, Manuel and Rota Bul{\`o}, Samuel and Richardt, Christian and Ramanan, Deva and Scherer, Sebastian and Kontschieder, Peter},
  journal = {arXiv preprint arXiv:2509.13414},
  year    = {2025}
}

@inproceedings{chen2025easi3r,
  title     = {{Easi3R}: Estimating Disentangled Motion from {DUSt3R} Without Training},
  author    = {Chen, Xingyu and Chen, Yue and Xiu, Yuliang and Geiger, Andreas and Chen, Anpei},
  booktitle = {Proceedings of the IEEE/CVF International Conference on Computer Vision (ICCV)},
  year      = {2025},
  note      = {arXiv:2503.24391}
}

@inproceedings{chen2026human3r,
  title     = {{Human3R}: Everyone Everywhere All at Once},
  author    = {Chen, Yue and Chen, Xingyu and Xue, Yuxuan and Chen, Anpei and Xiu, Yuliang and Pons-Moll, Gerard},
  booktitle = {International Conference on Learning Representations (ICLR)},
  year      = {2026},
  note      = {arXiv:2510.06219}
}

@inproceedings{cabon2025must3r,
  title     = {{MUSt3R}: Multi-View Network for Stereo {3D} Reconstruction},
  author    = {Cabon, Yohann and Stoffl, Lucas and Antsfeld, Leonid and Csurka, Gabriela and Chidlovskii, Boris and Revaud, J{\'e}r{\^o}me and Leroy, Vincent},
  booktitle = {Proceedings of the IEEE/CVF Conference on Computer Vision and Pattern Recognition (CVPR)},
  year      = {2025}
}

@inproceedings{jang2025pow3r,
  title     = {{Pow3R}: Empowering Unconstrained {3D} Reconstruction with Camera and Scene Priors},
  author    = {Jang, Wonbong and Weinzaepfel, Philippe and Leroy, Vincent and Agapito, Lourdes and Revaud, J{\'e}r{\^o}me},
  booktitle = {Proceedings of the IEEE/CVF Conference on Computer Vision and Pattern Recognition (CVPR)},
  pages     = {1071--1081},
  year      = {2025}
}

@inproceedings{zhang2025flare,
  title     = {{FLARE}: Feed-forward Geometry, Appearance and Camera Estimation from Uncalibrated Sparse Views},
  author    = {Zhang, Shangzhan and Wang, Jianyuan and Xu, Yinghao and Xue, Nan and Rupprecht, Christian and Zhou, Xiaowei and Shen, Yujun and Wetzstein, Gordon},
  booktitle = {Proceedings of the IEEE/CVF Conference on Computer Vision and Pattern Recognition (CVPR)},
  pages     = {21936--21947},
  year      = {2025}
}

@article{wang2025amb3r,
  title   = {{AMB3R}: Accurate Feed-forward Metric-scale {3D} Reconstruction with Backend},
  author  = {Wang, Hengyi and Agapito, Lourdes},
  journal = {arXiv preprint arXiv:2511.20343},
  year    = {2025}
}

@article{oquab2024dinov2,
  title   = {{DINOv2}: Learning Robust Visual Features without Supervision},
  author  = {Oquab, Maxime and Darcet, Timoth{\'e}e and Moutakanni, Th{\'e}o and Vo, Huy and Szafraniec, Marc and Khalidov, Vasil and Fernandez, Pierre and Haziza, Daniel and Massa, Francisco and El-Nouby, Alaaeldin and Assran, Mahmoud and Ballas, Nicolas and Galuba, Wojciech and Howes, Russell and Huang, Po-Yao and Li, Shang-Wen and Misra, Ishan and Rabbat, Michael and Sharma, Vasu and Synnaeve, Gabriel and Xu, Hu and Jegou, Herv{\'e} and Mairal, Julien and Labatut, Patrick and Joulin, Armand and Bojanowski, Piotr},
  journal = {Transactions on Machine Learning Research (TMLR)},
  year    = {2024}
}

@article{dong2024flexattention,
  title   = {Flex Attention: A Programming Model for Generating Optimized Attention Kernels},
  author  = {Dong, Juechu and Feng, Boyuan and Guessous, Driss and Liang, Yanbo and He, Horace},
  journal = {arXiv preprint arXiv:2412.05496},
  year    = {2024}
}

@inproceedings{wang2025spann3r,
  title     = {{3D} Reconstruction with Spatial Memory},
  author    = {Wang, Hengyi and Agapito, Lourdes},
  booktitle = {International Conference on {3D} Vision ({3DV})},
  year      = {2025}
}

@inproceedings{wang2025cut3r,
  title     = {Continuous {3D} Perception Model with Persistent State},
  author    = {Wang, Qianqian and Zhang, Yifei and Holynski, Aleksander and Efros, Alexei A. and Kanazawa, Angjoo},
  booktitle = {Proceedings of the IEEE/CVF Conference on Computer Vision and Pattern Recognition (CVPR)},
  year      = {2025}
}

@inproceedings{wu2025point3r,
  title     = {{Point3R}: Streaming {3D} Reconstruction with Explicit Spatial Pointer Memory},
  author    = {Wu, Yuqi and Zheng, Wenzhao and Zhou, Jie and Lu, Jiwen},
  booktitle = {Advances in Neural Information Processing Systems (NeurIPS)},
  year      = {2025}
}

@article{lan2025stream3r,
  title   = {{STream3R}: Scalable Sequential {3D} Reconstruction with Causal Transformer},
  author  = {Lan, Yushi and Luo, Yihang and Hong, Fangzhou and Zhou, Shangchen and Chen, Honghua and Lyu, Zhaoyang and Yang, Shuai and Dai, Bo and Loy, Chen Change and Pan, Xingang},
  journal = {arXiv preprint arXiv:2508.10893},
  year    = {2025}
}

@inproceedings{zhuo2026streamvggt,
  title     = {Streaming {4D} Visual Geometry Transformer},
  author    = {Zhuo, Dong and Zheng, Wenzhao and Guo, Jiahe and Wu, Yuqi and Zhou, Jie and Lu, Jiwen},
  booktitle = {International Conference on Learning Representations (ICLR)},
  year      = {2026},
  note      = {arXiv:2507.11539}
}

@inproceedings{chen2025long3r,
  title     = {{LONG3R}: Long Sequence Streaming {3D} Reconstruction},
  author    = {Chen, Zhuoguang and Qin, Minghui and Yuan, Tianyuan and Liu, Zhe and Zhao, Hang},
  booktitle = {Proceedings of the IEEE/CVF International Conference on Computer Vision (ICCV)},
  year      = {2025},
  note      = {arXiv:2507.18255}
}

@inproceedings{li2026wint3r,
  title     = {{WinT3R}: Window-Based Streaming Reconstruction with Camera Token Pool},
  author    = {Li, Zizun and Zhou, Jianjun and Wang, Yifan and Guo, Haoyu and Chang, Wenzheng and Zhou, Yang and Zhu, Haoyi and Chen, Junyi and Shen, Chunhua and He, Tong},
  booktitle = {International Conference on Learning Representations (ICLR)},
  year      = {2026},
  note      = {arXiv:2509.05296}
}

@article{cheng2026longstream,
  title   = {{LongStream}: Long-Sequence Streaming Autoregressive Visual Geometry},
  author  = {Cheng, Chong and Chen, Xianda and Xie, Tao and Yin, Wei and Ren, Weiqiang and Zhang, Qian and Guo, Xiaoyang and Wang, Hao},
  journal = {arXiv preprint arXiv:2602.13172},
  year    = {2026}
}

@article{chen2025ttt3r,
  title={{TTT3R}: {3D} Reconstruction as Test-Time Training},
  author={Chen, Xingyu and Chen, Yue and Xiu, Yuliang and Geiger, Andreas and Chen, Anpei},
  journal={arXiv preprint arXiv:2509.26645},
  year={2025}
}

@article{elflein2026vggt3,
  title   = {{VGG-T$^3$}: Offline Feed-Forward {3D} Reconstruction at Scale},
  author  = {Elflein, Sven and Li, Ruilong and Agostinho, S{\'e}rgio and Gojcic, Zan and Leal-Taix{\'e}, Laura and Zhou, Qunjie and Osep, Aljosa},
  journal = {arXiv preprint arXiv:2602.23361},
  year    = {2026}
}

@inproceedings{shen2025fastvggt,
  title     = {{FastVGGT}: Training-Free Acceleration of Visual Geometry Transformer},
  author    = {Shen, You and Zhang, Zhipeng and Qu, Yansong and Zheng, Xiawu and Ji, Jiayi and Zhang, Shengchuan and Cao, Liujuan},
  booktitle = {International Conference on Learning Representations (ICLR)},
  year      = {2026},
  note      = {arXiv:2509.02560}
}

@inproceedings{jin2026zipmap,
  title     = {{ZipMap}: Linear-Time Stateful {3D} Reconstruction via Test-Time Training},
  author    = {Jin, Haian and Wu, Rundi and Zhang, Tianyuan and Gao, Ruiqi and Barron, Jonathan T. and Snavely, Noah and Ho{\l}y{\'n}ski, Aleksander},
  booktitle = {Proceedings of the IEEE/CVF Conference on Computer Vision and Pattern Recognition (CVPR)},
  year      = {2026},
  note      = {arXiv:2603.04385}
}

@inproceedings{liu2025slam3r,
  title     = {{SLAM3R}: Real-Time Dense Scene Reconstruction from Monocular {RGB} Videos},
  author    = {Liu, Yuzheng and Dong, Siyan and Wang, Shuzhe and Yin, Yingda and Yang, Yanchao and Fan, Qingnan and Chen, Baoquan},
  booktitle = {Proceedings of the IEEE/CVF Conference on Computer Vision and Pattern Recognition (CVPR)},
  year      = {2025},
  note      = {arXiv:2412.09401}
}

@inproceedings{zhang2025monst3r,
  title     = {{MonST3R}: A Simple Approach for Estimating Geometry in the Presence of Motion},
  author    = {Zhang, Junyi and Herrmann, Charles and Hur, Junhwa and Jampani, Varun and Darrell, Trevor and Cole, Forrester and Sun, Deqing and Yang, Ming-Hsuan},
  booktitle = {International Conference on Learning Representations (ICLR)},
  year      = {2025},
  note      = {arXiv:2410.03825}
}

@article{yuan2026infinitevggt,
  title   = {{InfiniteVGGT}: Visual Geometry Grounded Transformer for Endless Streams},
  author  = {Yuan, Shuai and Yang, Yantai and Yang, Xiaotian and Zhang, Xupeng and Zhao, Zhonghao and Zhang, Lingming and Zhang, Zhipeng},
  journal = {arXiv preprint arXiv:2601.02281},
  year    = {2026}
}

@article{zhang2026loger,
  title   = {{LoGeR}: Long-Context Geometric Reconstruction with Hybrid Memory},
  author  = {Zhang, Junyi and Herrmann, Charles and Hur, Junhwa and Sun, Chen and Yang, Ming-Hsuan and Cole, Forrester and Darrell, Trevor and Sun, Deqing},
  journal = {arXiv preprint arXiv:2603.03269},
  year    = {2026}
}

@article{lu2026ovggt,
  title   = {{OVGGT}: {O(1)} Constant-Cost Streaming Visual Geometry Transformer},
  author  = {Lu, Si-Yu and Chen, Po-Ting and Hsu, Hui-Che and Jhong, Sin-Ye and Cheng, Wen-Huang and Chen, Yung-Yao},
  journal = {arXiv preprint arXiv:2603.05959},
  year    = {2026}
}

@article{xie2026scal3r,
  title   = {{Scal3R}: Scalable Test-Time Training for Large-Scale {3D} Reconstruction},
  author  = {Xie, Tao and Yang, Peishan and Jin, Yudong and Cai, Yingfeng and Yin, Wei and Ren, Weiqiang and Zhang, Qian and Hua, Wei and Peng, Sida and Guo, Xiaoyang and Zhou, Xiaowei},
  journal = {arXiv preprint arXiv:2604.08542},
  year    = {2026}
}

@article{chen2026lingbotmap,
  title   = {Geometric Context Transformer for Streaming {3D} Reconstruction},
  author  = {Chen, Lin-Zhuo and Gao, Jian and Chen, Yihang and Cheng, Ka Leong and Sun, Yipengjing and Hu, Liangxiao and Xue, Nan and Zhu, Xing and Shen, Yujun and Yao, Yao and Xu, Yinghao},
  journal = {arXiv preprint arXiv:2604.14141},
  year    = {2026}
}

@article{taher2025kvtracker,
  title   = {{KV-Tracker}: Real-Time Pose Tracking with Transformers},
  author  = {Taher, Marwan and Alzugaray, Ignacio and Mazur, Kirill and Kong, Xin and Davison, Andrew J.},
  journal = {arXiv preprint arXiv:2512.22581},
  year    = {2025}
}

@inproceedings{schonberger2016structure,
  title     = {Structure-from-Motion Revisited},
  author    = {Sch{\"o}nberger, Johannes Lutz and Frahm, Jan-Michael},
  booktitle = {Proceedings of the IEEE Conference on Computer Vision and Pattern Recognition (CVPR)},
  year      = {2016}
}

@inproceedings{pan2024glomap,
  title     = {Global Structure-from-Motion Revisited},
  author    = {Pan, Linfei and Barath, Daniel and Pollefeys, Marc and Sch{\"o}nberger, Johannes Lutz},
  booktitle = {European Conference on Computer Vision (ECCV)},
  year      = {2024}
}

@article{murartal2017orbslam2,
  title   = {{ORB-SLAM2}: An Open-Source {SLAM} System for Monocular, Stereo, and {RGB-D} Cameras},
  author  = {Mur-Artal, Ra{\'u}l and Tard{\'o}s, Juan D.},
  journal = {IEEE Transactions on Robotics},
  volume  = {33},
  number  = {5},
  pages   = {1255--1262},
  year    = {2017}
}

@article{campos2021orbslam3,
  title   = {{ORB-SLAM3}: An Accurate Open-Source Library for Visual, Visual-Inertial, and Multimap {SLAM}},
  author  = {Campos, Carlos and Elvira, Richard and Rodr{\'i}guez, Juan J. G{\'o}mez and Montiel, J. M. M. and Tard{\'o}s, Juan D.},
  journal = {IEEE Transactions on Robotics},
  volume  = {37},
  number  = {6},
  pages   = {1874--1890},
  year    = {2021}
}

@article{furukawa2010accurate,
  title   = {Accurate, Dense, and Robust Multi-View Stereopsis},
  author  = {Furukawa, Yasutaka and Ponce, Jean},
  journal = {IEEE Transactions on Pattern Analysis and Machine Intelligence (TPAMI)},
  volume  = {32},
  number  = {8},
  pages   = {1362--1376},
  year    = {2010}
}

@inproceedings{schonberger2016pixelwise,
  title     = {Pixelwise View Selection for Unstructured Multi-View Stereo},
  author    = {Sch{\"o}nberger, Johannes Lutz and Zheng, Enliang and Frahm, Jan-Michael and Pollefeys, Marc},
  booktitle = {European Conference on Computer Vision (ECCV)},
  year      = {2016}
}

@inproceedings{yao2018mvsnet,
  title     = {{MVSNet}: Depth Inference for Unstructured Multi-View Stereo},
  author    = {Yao, Yao and Luo, Zixin and Li, Shiwei and Fang, Tian and Quan, Long},
  booktitle = {European Conference on Computer Vision (ECCV)},
  year      = {2018}
}

@inproceedings{detone2018superpoint,
  title     = {{SuperPoint}: Self-Supervised Interest Point Detection and Description},
  author    = {DeTone, Daniel and Malisiewicz, Tomasz and Rabinovich, Andrew},
  booktitle = {Proceedings of the IEEE Conference on Computer Vision and Pattern Recognition Workshops (CVPRW)},
  year      = {2018}
}

@inproceedings{sarlin2020superglue,
  title     = {{SuperGlue}: Learning Feature Matching with Graph Neural Networks},
  author    = {Sarlin, Paul-Edouard and DeTone, Daniel and Malisiewicz, Tomasz and Rabinovich, Andrew},
  booktitle = {Proceedings of the IEEE/CVF Conference on Computer Vision and Pattern Recognition (CVPR)},
  year      = {2020}
}

@inproceedings{lindenberger2023lightglue,
  title     = {{LightGlue}: Local Feature Matching at Light Speed},
  author    = {Lindenberger, Philipp and Sarlin, Paul-Edouard and Pollefeys, Marc},
  booktitle = {Proceedings of the IEEE/CVF International Conference on Computer Vision (ICCV)},
  pages     = {17627--17638},
  year      = {2023}
}

@inproceedings{sun2021loftr,
  title     = {{LoFTR}: Detector-Free Local Feature Matching with Transformers},
  author    = {Sun, Jiaming and Shen, Zehong and Wang, Yuang and Bao, Hujun and Zhou, Xiaowei},
  booktitle = {Proceedings of the IEEE/CVF Conference on Computer Vision and Pattern Recognition (CVPR)},
  year      = {2021}
}

@inproceedings{wang2024eloftr,
  title     = {{Efficient LoFTR}: Semi-Dense Local Feature Matching with Sparse-Like Speed},
  author    = {Wang, Yifan and He, Xingyi and Peng, Sida and Tan, Dongli and Zhou, Xiaowei},
  booktitle = {Proceedings of the IEEE/CVF Conference on Computer Vision and Pattern Recognition (CVPR)},
  year      = {2024}
}

@inproceedings{wang2024vggsfm,
  title     = {{VGGSfM}: Visual Geometry Grounded Deep Structure From Motion},
  author    = {Wang, Jianyuan and Karaev, Nikita and Rupprecht, Christian and Novotny, David},
  booktitle = {Proceedings of the IEEE/CVF Conference on Computer Vision and Pattern Recognition (CVPR)},
  year      = {2024}
}

@inproceedings{duisterhof2025mast3rsfm,
  title     = {{MASt3R-SfM}: A Fully-Integrated Solution for Unconstrained Structure-from-Motion},
  author    = {Duisterhof, Bardienus Pieter and Zust, Lojze and Weinzaepfel, Philippe and Leroy, Vincent and Cabon, Yohann and Revaud, J{\'e}r{\^o}me},
  booktitle = {International Conference on {3D} Vision ({3DV})},
  year      = {2025}
}

@inproceedings{murai2025mast3rslam,
  title     = {{MASt3R-SLAM}: Real-Time Dense {SLAM} with {3D} Reconstruction Priors},
  author    = {Murai, Riku and Dexheimer, Eric and Davison, Andrew J.},
  booktitle = {Proceedings of the IEEE/CVF Conference on Computer Vision and Pattern Recognition (CVPR)},
  year      = {2025},
  note      = {arXiv:2412.12392}
}

@article{maggio2025vggtslam,
  title   = {{VGGT-SLAM}: Dense {RGB} {SLAM} Optimized on the {SL(4)} Manifold},
  author  = {Maggio, Dominic and Lim, Hyungtae and Carlone, Luca},
  journal = {arXiv preprint arXiv:2505.12549},
  year    = {2025}
}

@inproceedings{teed2021droidslam,
  title     = {{DROID-SLAM}: Deep Visual {SLAM} for Monocular, Stereo, and {RGB-D} Cameras},
  author    = {Teed, Zachary and Deng, Jia},
  booktitle = {Advances in Neural Information Processing Systems (NeurIPS)},
  year      = {2021}
}

@inproceedings{teed2023dpvo,
  title     = {Deep Patch Visual Odometry},
  author    = {Teed, Zachary and Lipson, Lahav and Deng, Jia},
  booktitle = {Advances in Neural Information Processing Systems (NeurIPS)},
  year      = {2023}
}

@inproceedings{li2025megasam,
  title     = {{MegaSaM}: Accurate, Fast, and Robust Structure and Motion from Casual Dynamic Videos},
  author    = {Li, Zhengqi and Tucker, Richard and Cole, Forrester and Wang, Qianqian and Jin, Linyi and Ye, Vickie and Kanazawa, Angjoo and Ho{\l}y{\'n}ski, Aleksander and Snavely, Noah},
  booktitle = {Proceedings of the IEEE/CVF Conference on Computer Vision and Pattern Recognition (CVPR)},
  pages     = {10486--10496},
  year      = {2025}
}

@inproceedings{klein2007ptam,
  title     = {Parallel Tracking and Mapping for Small {AR} Workspaces},
  author    = {Klein, Georg and Murray, David},
  booktitle = {IEEE and ACM International Symposium on Mixed and Augmented Reality (ISMAR)},
  year      = {2007}
}

@article{murartal2015orbslam,
  title   = {{ORB-SLAM}: A Versatile and Accurate Monocular {SLAM} System},
  author  = {Mur-Artal, Ra{\'u}l and Montiel, J. M. M. and Tard{\'o}s, Juan D.},
  journal = {IEEE Transactions on Robotics},
  volume  = {31},
  number  = {5},
  pages   = {1147--1163},
  year    = {2015}
}

@article{engel2017dso,
  title   = {Direct Sparse Odometry},
  author  = {Engel, Jakob and Koltun, Vladlen and Cremers, Daniel},
  journal = {IEEE Transactions on Pattern Analysis and Machine Intelligence (TPAMI)},
  volume  = {40},
  number  = {3},
  pages   = {611--625},
  year    = {2018}
}

@inproceedings{butler2012sintel,
  title     = {A Naturalistic Open Source Movie for Optical Flow Evaluation},
  author    = {Butler, Daniel J. and Wulff, Jonas and Stanley, Garrett B. and Black, Michael J.},
  booktitle = {European Conference on Computer Vision (ECCV)},
  year      = {2012}
}

@inproceedings{sturm2012tum,
  title     = {A Benchmark for the Evaluation of {RGB-D} {SLAM} Systems},
  author    = {Sturm, J{\"u}rgen and Engelhard, Nikolas and Endres, Felix and Burgard, Wolfram and Cremers, Daniel},
  booktitle = {IEEE/RSJ International Conference on Intelligent Robots and Systems (IROS)},
  year      = {2012}
}

@inproceedings{dai2017scannet,
  title     = {{ScanNet}: Richly-Annotated {3D} Reconstructions of Indoor Scenes},
  author    = {Dai, Angela and Chang, Angel X. and Savva, Manolis and Halber, Maciej and Funkhouser, Thomas and Nie{\ss}ner, Matthias},
  booktitle = {Proceedings of the IEEE Conference on Computer Vision and Pattern Recognition (CVPR)},
  year      = {2017}
}

@inproceedings{shotton2013scenecoord,
  title     = {Scene Coordinate Regression Forests for Camera Relocalization in {RGB-D} Images},
  author    = {Shotton, Jamie and Glocker, Ben and Zach, Christopher and Izadi, Shahram and Criminisi, Antonio and Fitzgibbon, Andrew},
  booktitle = {Proceedings of the IEEE Conference on Computer Vision and Pattern Recognition (CVPR)},
  year      = {2013}
}

@InProceedings{azinovic2022neuralrgbd,
    author    = {Azinovi\'c, Dejan and Martin-Brualla, Ricardo and Goldman, Dan B and Nie{\ss}ner, Matthias and Thies, Justus},
    title     = {Neural RGB-D Surface Reconstruction},
    booktitle = {Proceedings of the IEEE/CVF Conference on Computer Vision and Pattern Recognition (CVPR)},
    month     = {June},
    year      = {2022},
    pages     = {6290-6301}
}

@inproceedings{palazzolo2019refusion,
  title     = {{ReFusion}: {3D} Reconstruction in Dynamic Environments for {RGB-D} Cameras Exploiting Residuals},
  author    = {Palazzolo, Emanuele and Behley, Jens and Lottes, Philipp and Gigu{\`e}re, Philippe and Stachniss, Cyrill},
  booktitle = {IEEE/RSJ International Conference on Intelligent Robots and Systems (IROS)},
  year      = {2019}
}

@inproceedings{geiger2012kitti,
  title     = {Are We Ready for Autonomous Driving? The {KITTI} Vision Benchmark Suite},
  author    = {Geiger, Andreas and Lenz, Philip and Urtasun, Raquel},
  booktitle = {Proceedings of the IEEE Conference on Computer Vision and Pattern Recognition (CVPR)},
  year      = {2012}
}

@inproceedings{reizenstein2021co3d,
  title     = {Common Objects in {3D}: Large-Scale Learning and Evaluation of Real-life {3D} Category Reconstruction},
  author    = {Reizenstein, Jeremy and Shapovalov, Roman and Henzler, Philipp and Sbordone, Luca and Labatut, Patrick and Novotny, David},
  booktitle = {IEEE/CVF International Conference on Computer Vision (ICCV)},
  year      = {2021}
}

@inproceedings{xia2024wildrgbd,
  title     = {{RGBD} Objects in the Wild: Scaling Real-World {3D} Object Learning from {RGB-D} Videos},
  author    = {Xia, Hongchi and Fu, Yang and Liu, Sifei and Wang, Xiaolong},
  booktitle = {Proceedings of the IEEE/CVF Conference on Computer Vision and Pattern Recognition (CVPR)},
  year      = {2024}
}

@inproceedings{baruch2021arkitscenes,
  title     = {{ARKitScenes}: A Diverse Real-World Dataset for {3D} Indoor Scene Understanding Using Mobile {RGB-D} Data},
  author    = {Baruch, Gilad and Chen, Zhuoyuan and Dehghan, Afshin and Dimry, Tal and Feigin, Yuri and Fu, Peter and Gebauer, Thomas and Joffe, Brandon and Kurz, Daniel and Schwartz, Arik and Shulman, Elad},
  booktitle = {Advances in Neural Information Processing Systems (NeurIPS) Datasets and Benchmarks Track},
  year      = {2021}
}

@inproceedings{yeshwanth2023scannetpp,
  title     = {{ScanNet++}: A High-Fidelity Dataset of {3D} Indoor Scenes},
  author    = {Yeshwanth, Chandan and Liu, Yueh-Cheng and Nie{\ss}ner, Matthias and Dai, Angela},
  booktitle = {IEEE/CVF International Conference on Computer Vision (ICCV)},
  year      = {2023}
}

@inproceedings{arnold2022mapfree,
  title     = {Map-free Visual Relocalization: Metric Pose Relative to a Single Image},
  author    = {Arnold, Eduardo and Wynn, Jamie and Vicente, Sara and Garcia-Hernando, Guillermo and Monszpart, {\'A}ron and Prisacariu, Victor and Turmukhambetov, Daniyar and Brachmann, Eric},
  booktitle = {European Conference on Computer Vision (ECCV)},
  year      = {2022}
}

@inproceedings{ling2024dl3dv,
  title     = {{DL3DV-10K}: A Large-Scale Scene Dataset for Deep Learning-based {3D} Vision},
  author    = {Ling, Lu and Sheng, Yichen and Tu, Zhi and Zhao, Wentian and Xin, Cheng and Wan, Kun and Yu, Lantao and Guo, Qianyu and Yu, Zixun and Lu, Yawen and Li, Xuanmao and Sun, Xingpeng and Ashok, Rohan and Mukherjee, Aniruddha and Kang, Hao and Kong, Xiangrui and Hua, Gang and Zhang, Tianyi and Benes, Bedrich and Bera, Aniket},
  booktitle = {Proceedings of the IEEE/CVF Conference on Computer Vision and Pattern Recognition (CVPR)},
  year      = {2024}
}

@inproceedings{wang2020tartanair,
  title     = {{TartanAir}: A Dataset to Push the Limits of Visual {SLAM}},
  author    = {Wang, Wenshan and Zhu, Delong and Wang, Xiangwei and Hu, Yaoyu and Qiu, Yuheng and Wang, Chen and Hu, Yafei and Kapoor, Ashish and Scherer, Sebastian},
  booktitle = {IEEE/RSJ International Conference on Intelligent Robots and Systems (IROS)},
  year      = {2020}
}

@inproceedings{mehl2023spring,
  title     = {Spring: A High-Resolution High-Detail Dataset and Benchmark for Scene Flow, Optical Flow and Stereo},
  author    = {Mehl, Lukas and Schmalfuss, Jenny and Jahedi, Azin and Nalivayko, Yaroslava and Bruhn, Andr{\'e}s},
  booktitle = {Proceedings of the IEEE/CVF Conference on Computer Vision and Pattern Recognition (CVPR)},
  year      = {2023}
}

@inproceedings{huang2018deepmvs,
  title     = {{DeepMVS}: Learning Multi-view Stereopsis},
  author    = {Huang, Po-Han and Matzen, Kevin and Kopf, Johannes and Ahuja, Narendra and Huang, Jia-Bin},
  booktitle = {Proceedings of the IEEE Conference on Computer Vision and Pattern Recognition (CVPR)},
  year      = {2018}
}

@inproceedings{roberts2021hypersim,
  title     = {Hypersim: A Photorealistic Synthetic Dataset for Holistic Indoor Scene Understanding},
  author    = {Roberts, Mike and Ramapuram, Jason and Ranjan, Anurag and Kumar, Atulit and Bautista, Miguel Angel and Paczan, Nathan and Webb, Russ and Susskind, Joshua M.},
  booktitle = {IEEE/CVF International Conference on Computer Vision (ICCV)},
  year      = {2021}
}

@article{zhou2025omniworld,
  title   = {{OmniWorld}: A Multi-Domain and Multi-Modal Dataset for {4D} World Modeling},
  author  = {Zhou, Yang and Wang, Yifan and Zhou, Jianjun and Chang, Wenzheng and Guo, Haoyu and Li, Zizun and Ma, Kaijing and Li, Xinyue and Wang, Yating and Zhu, Haoyi and Liu, Mingyu and Liu, Dingning and Yang, Jiange and Fu, Zhoujie and Chen, Junyi and Shen, Chunhua and Pang, Jiangmiao and Zhang, Kaipeng and He, Tong},
  journal = {arXiv preprint arXiv:2509.12201},
  year    = {2025}
}

@article{cabon2020vkitti2,
  title   = {Virtual {KITTI} 2},
  author  = {Cabon, Yohann and Murray, Naila and Humenberger, Martin},
  journal = {arXiv preprint arXiv:2001.10773},
  year    = {2020}
}

@inproceedings{karaev2023dynamicreplica,
  title     = {{DynamicStereo}: Consistent Dynamic Depth from Stereo Videos},
  author    = {Karaev, Nikita and Rocco, Ignacio and Graham, Benjamin and Neverova, Natalia and Vedaldi, Andrea and Rupprecht, Christian},
  booktitle = {Proceedings of the IEEE/CVF Conference on Computer Vision and Pattern Recognition (CVPR)},
  year      = {2023}
}

@article{han2025emergent,
  title   = {Emergent Outlier View Rejection in Visual Geometry Grounded Transformers},
  author  = {Han, Jisang and Hong, Sunghwan and Jung, Jaewoo and Jang, Wooseok and An, Honggyu and Wang, Qianqian and Kim, Seungryong and Feng, Chen},
  journal = {arXiv preprint arXiv:2512.04012},
  year    = {2025}
}

@inproceedings{schops2017eth3d,
  title     = {A Multi-View Stereo Benchmark with High-Resolution Images and Multi-Camera Videos},
  author    = {Sch{\"o}ps, Thomas and Sch{\"o}nberger, Johannes L. and Galliani, Silvano and Sattler, Torsten and Schindler, Konrad and Pollefeys, Marc and Geiger, Andreas},
  booktitle = {Proceedings of the IEEE Conference on Computer Vision and Pattern Recognition (CVPR)},
  year      = {2017}
}

@inproceedings{sabour2023robustnerf,
  title     = {{RobustNeRF}: Ignoring Distractors with Robust Losses},
  author    = {Sabour, Sara and Vora, Suhani and Duckworth, Daniel and Krasin, Ivan and Fleet, David J. and Tagliasacchi, Andrea},
  booktitle = {Proceedings of the IEEE/CVF Conference on Computer Vision and Pattern Recognition (CVPR)},
  pages     = {20626--20636},
  month     = {June},
  year      = {2023}
}

@misc{aria_ase,
  title        = {Aria Synthetic Environments Dataset},
  author       = {{Project Aria}},
  howpublished = {\url{https://www.projectaria.com/datasets/ase/}},
  year         = {2024},
  note         = {Meta Reality Labs Research}
}

\clearpage
\appendix

\section{Reconstruction Gallery}\label{app:gallery}

\begin{figure}[H]
  \centering
  \includegraphics[width=\linewidth,height=1\textheight,keepaspectratio]{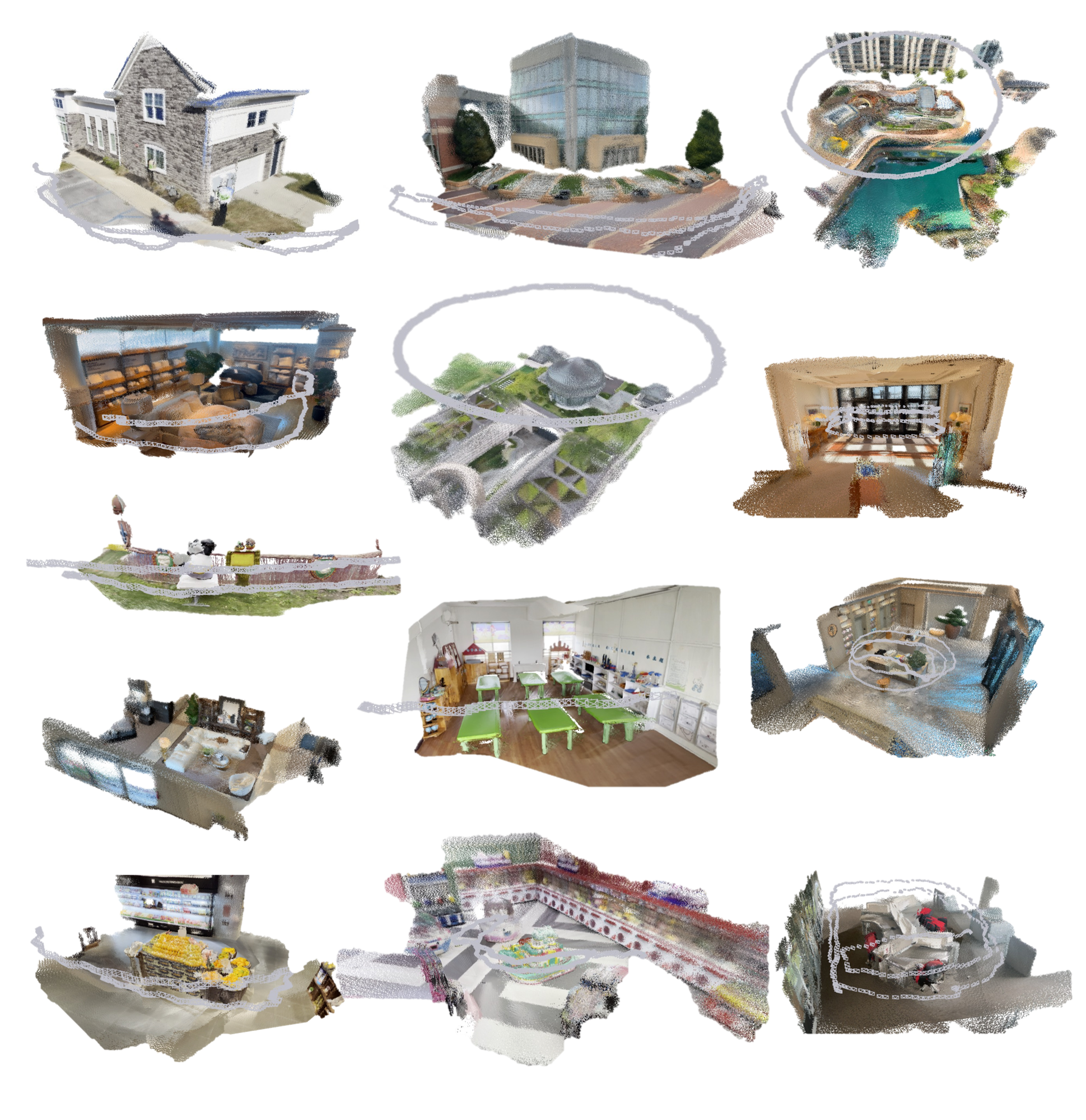}
\caption{\textbf{Reconstruction gallery.} Qualitative reconstruction results from \modelname in streaming mode across diverse indoor and outdoor scenes. The reconstructed point clouds remain geometrically coherent and visually consistent across varied scene layouts, object scales, and camera trajectories, demonstrating \modelname's ability to maintain stable scene structure during online reconstruction of long sequences.}
  \label{fig:reconstruction_gallery}
\end{figure}
\clearpage

\section{Long-Sequence Compute Cost}\label{app:compute_cost}

\begin{figure}[!ht]
    \vspace{-6pt}
  \centering
  \begin{subfigure}[b]{0.44\linewidth}
    \centering
    \includegraphics[width=\linewidth]{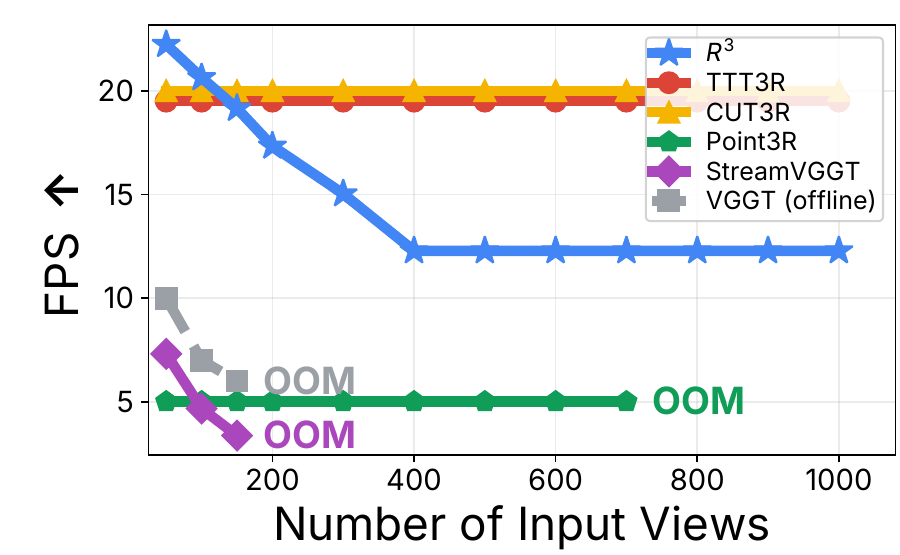}
    \vspace{-14pt}
  \end{subfigure}\hspace{6pt}%
  \begin{subfigure}[b]{0.44\linewidth}
    \centering
    \includegraphics[width=\linewidth]{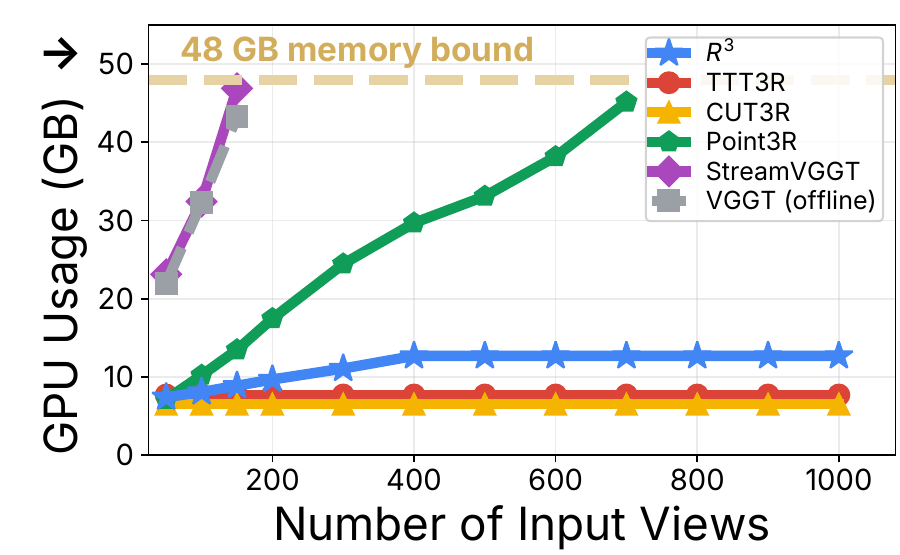}
    \vspace{-14pt}
  \end{subfigure}%
  \vspace{-4pt}
  \caption{\textbf{Long-sequence compute cost.} Inference FPS (left) and GPU memory usage (right) as the number of input views increases. ``OOM'' denotes out-of-memory.}
  \label{fig:compute_cost}
   \vspace{-10pt}
\end{figure}

\figref{fig:compute_cost} reports inference FPS and GPU memory usage under the same 7-Scenes protocol used in \secref{sec:longseq}. Global-regression baselines (e.g., StreamVGGT) either hit OOM or slow sharply as $N$ grows; \modelname replaces this $O(N^2)$ growth with a bounded memory increase and a gentler FPS decline, consistent with the bounded keyframe bank.

\section{Streaming Details and Optional Pose-Graph Refinement}\label{app:online_details}

This appendix provides additional details on the streaming machinery of \modelname: its connection to classical keyframe selection, the train/inference length-gap argument, outlier rejection, optional pose-graph refinement (used as an end-of-sequence cleanup rather than in the main-body numbers), and segment resets for long streams. The core learned mechanism---the confidence-driven keyframe bank---is described in the main text (\secref{sec:online}); the components below are auxiliary and follow common practice.

\paragraph{Comparison to classical keyframe selection.}
$\mathcal{B}_t$ plays the role of a \emph{keyframe database} in classical SLAM systems such as PTAM~\cite{klein2007ptam}, ORB-SLAM~\cite{murartal2015orbslam,murartal2017orbslam2,campos2021orbslam3}, and DSO~\cite{engel2017dso}: it is a compact subset of frames carried through the pipeline instead of the full stream. Classical systems trigger keyframe insertion using hand-tuned translation, rotation, or covisibility thresholds. Our admission rule replaces these cues with a learned-feature similarity test plus a maximum-staleness cap, leveraging the fact that the DA3~\cite{lin2025da3} frame-only encoder is pretrained on large-scale multi-view geometry, so its averaged frame token already reflects both visual content and scene-level geometry.

\paragraph{Train/inference length gap.}
The novelty gate keeps the active context informative without increasing memory unboundedly and slows long-term saturation, helping bridge the train/inference length gap: we train on sequences of at most $32$ frames but deploy on streams of hundreds. A strict novelty threshold keeps the effective context length close to the training regime even as wall-clock sequence length grows. To prevent staleness when scene content evolves slowly and no incoming frame is novel enough to pass the threshold, we force-admit a frame whenever no keyframe has been added in the last $\Delta_{\max}=20$ incoming frames. The culling cap is $M_{\max}=100$.
When this cap is exceeded, we evict the keyframe with the lowest utility
\[
u_j=d_j c_j,\qquad
d_j=\min_{i\in\mathcal{B}_t\setminus\{j\}}\!\big(1-\cos(\mathbf{tok}_i,\mathbf{tok}_j)\big),\qquad
c_j=\max_{i\in\mathcal{B}_t\setminus\{j\}}\!\tfrac{1}{2}\!\big(c^{\mathrm{R}}_{i\rightarrow j}+c^{\mathrm{T}}_{i\rightarrow j}\big).
\]
Here $d_j$ is small for frames that are redundant with another bank entry, and $c_j$ is small when the frame has no strong pose edge to the bank. The first frame is excluded from eviction.

\paragraph{Outlier rejection and segment reset.}
Let $\bar{c}_j=\tfrac{1}{|\mathcal{C}_t|}\sum_{i\in\mathcal{C}_t}\bar{c}_{i\rightarrow j}$ denote the mean averaged-pair confidence of frame $j$ against its current context. We calibrate a baseline $\bar{c}^{\mathrm{ref}}=\tfrac{1}{N_{\mathrm{cal}}}\sum_{j=1}^{N_{\mathrm{cal}}}\bar{c}_j$ from the first $N_{\mathrm{cal}}$ frames of the stream and reject any subsequent frame whose mean falls below $\tau_{\mathrm{out}}\,\bar{c}^{\mathrm{ref}}$. This check is post-hoc: rejected frames have already passed through the backbone because their tokens are needed to score the pair head. Upon rejection, however, their entries are evicted from the backbone KV cache so they do not pollute later causal attention; they also do not enter $\mathcal{B}_t$, contribute no edges to $\mathcal{E}_t$, or receive an estimated absolute pose. The threshold is loose enough to retain difficult but recoverable frames while filtering catastrophic cases such as severe motion blur, near-total occlusion, or scene cuts mid-frame. The same test also serves as a tracking-failure detector: after $N_{\mathrm{rej}}$ consecutive rejected frames, we declare loss of track and trigger a segment reset (below) rather than continue with a degraded context. The keyframe-bank cap $M_{\max}$ is a secondary bound that rarely fires before this confidence trigger in practice. We use $N_{\mathrm{cal}}=3$, $\tau_{\mathrm{out}}=0.15$, and $N_{\mathrm{rej}}=3$ in our experiments.

\paragraph{Optional pose-graph refinement for streaming runs.}
The full-context mode (\secref{sec:inference}) closes with a confidence-weighted pose-graph optimization over the full pair edge set. The same step can be invoked as an end-of-stream refinement on a streaming run --- using the streaming trajectory as initialization and the edge set $\mathcal{E}$ restricted to pairs that were evaluated against the keyframe bank during the stream --- or at every segment boundary for jobs that tolerate occasional latency spikes in exchange for tighter mid-stream geometry. We do not apply this refinement to any reported streaming main-body number; it appears only as an ablation.
Given an edge set $\mathcal{E}$ and an aggregated initialization $\{\mathbf{T}_i\}$, let $e^{\mathrm{R}}_{ij}$ and $e^{\mathrm{T}}_{ij}$ denote the rotation and translation residuals between the relative pose induced by $(\mathbf{T}_i,\mathbf{T}_j)$ and the predicted edge $(\hat{\mathbf{q}}_{i\rightarrow j},\hat{\mathbf{t}}_{i\rightarrow j})$. We solve
\[
\min_{\{\mathbf{T}_i\}}\sum_{(i,j)\in\mathcal{E}}\Big(c^{\mathrm{R}}_{i\rightarrow j}\,H_{\delta_{\mathrm{R}}}(e^{\mathrm{R}}_{ij})+c^{\mathrm{T}}_{i\rightarrow j}\,H_{\delta_{\mathrm{T}}}(e^{\mathrm{T}}_{ij})\Big),
\]
with Huber losses $H_{\delta_{\mathrm{R}}}, H_{\delta_{\mathrm{T}}}$, $\mathbf{T}_1$ fixed, and L-BFGS as the solver. This is a lightweight pose-only refinement: edges carry only relative-pose residuals and scalar confidences, with no bundle adjustment, point reprojection, or depth optimization.

\paragraph{Segment reset and bridges.}
A segment ends either when sustained low confidence triggers the reset above or when the sequence length cap $L_{\max}$ is reached on a very long stream. In both cases, we link consecutive segments with a short \emph{bridge} of $3$--$10$ shared frames. When advancing to a new segment, we clear the DA3 cache and keyframe bank, then rerun the backbone on the bridge frames together with the incoming stream.

Global pose continuity follows by composition: each bridge frame $b$ carries an absolute pose $\mathbf{T}^{\mathrm{abs}}_{b}$ from the previous segment, and for each new frame $j$ we recover $\mathbf{T}^{\mathrm{abs}}_{j}=\mathbf{T}^{\mathrm{abs}}_{b}\cdot\mathbf{T}_{b\rightarrow j}$, without requiring Sim(3) or $\mathrm{SE}(3)$ registration across segments.

Scale can drift between independent forward passes. We anchor it by aligning the DA3 depth prediction on each segment's first frame to a pretrained metric-depth model with a single median-based scalar, which is applied uniformly to all pose translations and depths in that segment. Per-boundary error is therefore reduced to one metric-anchored scalar rather than the joint rotation, translation, and scale drift of Sim(3)-style window fusion.

\section{Pose Supervision and Learned Pair Reliability}\label{app:pose_supervision}

\subsection{Pose Loss Formulations}\label{app:pose_loss_formulations}
\figref{fig:method_comparison} compares three feed-forward pose paradigms; here we spell out their corresponding supervision targets. Let $\mathbf{T}_i=(\mathbf{q}_i,\mathbf{t}_i)$ denote the camera-to-world pose of frame $i$, and let $\mathbf{T}_{i\rightarrow j}=\mathbf{T}_i^{-1}\mathbf{T}_j$ denote the relative transform from frame $i$ to frame $j$.

VGGT-style training~\cite{wang2025vggt} supervises poses directly in one canonical coordinate system. Abstractly, with frame~$1$ used as the global anchor, the loss takes the form
\[
\mathcal{L}_{\mathrm{abs}}
=\sum_{j=1}^{N} d_{\mathrm{pose}}\!\left(\hat{\mathbf{T}}_j,\mathbf{T}^{*}_j\right),
\]
where all targets $\mathbf{T}^{*}_j$ are expressed in the same anchored world frame. This target is simple, but the network must learn a coordinate choice that is arbitrary from the scene's perspective and increasingly difficult to preserve as the trajectory grows.

$\pi^3$~\cite{wang2026pi3} moves away from VGGT's fixed first-frame coordinate target: it predicts one pose per view and supervises them through all-pair relative-pose losses:
\[
\mathcal{L}_{\pi^3}
=\sum_{i\neq j}
d_{\mathrm{pose}}\!\left(\hat{\mathbf{T}}_i^{-1}\hat{\mathbf{T}}_j,\mathbf{T}^{*}_{i\rightarrow j}\right).
\]
This removes the need to match a fixed first-frame coordinate system at the loss level and provides $O(N^2)$ pairwise constraints. However, it is still not ideal for online reconstruction. The all-pair loss is defined over a complete set of views, whereas a causal stream must emit estimates from prefixes and cannot use future observations unless it revises past outputs. In addition, every pair contributes with the same weight, so difficult or weakly constrained pairs can consume the same gradient budget as reliable ones.

Our formulation predicts each relative transform directly and attaches separate learned confidences to rotation and translation:
\[
\left(\hat{\mathbf{q}}_{i\rightarrow j},\hat{\mathbf{t}}_{i\rightarrow j},
c^{\mathrm{R}}_{i\rightarrow j},c^{\mathrm{T}}_{i\rightarrow j}\right)
=\mathrm{MLP}_{\mathrm{rel}}\!\left([\mathbf{z}_i;\mathbf{z}_j]\right).
\]
The pose loss uses confidence-weighted residuals. Omitting the per-frame focal-length term for clarity, we define
\[
\mathcal{L}_{\mathrm{rot}}(i,j)=c^{\mathrm{R}}_{i\rightarrow j}\,\ell^{L_1}_{\mathrm{rot}}(i,j)-\alpha\log c^{\mathrm{R}}_{i\rightarrow j},
\qquad
\mathcal{L}_{\mathrm{trans}}(i,j)=c^{\mathrm{T}}_{i\rightarrow j}\,\ell^{L_1}_{\mathrm{trans}}(i,j)-\alpha\log c^{\mathrm{T}}_{i\rightarrow j},
\]
and average over a set of ordered pairs,
\[
\mathcal{L}_{\mathrm{cam}}
=\frac{1}{|\mathcal{P}|}
\sum_{(i,j)\in\mathcal{P}}\Big(
\mathcal{L}_{\mathrm{rot}}(i,j)+\mathcal{L}_{\mathrm{trans}}(i,j)
\Big).
\]
For the causal checkpoint, $\mathcal{P}$ is the past-to-current lower-triangular pair set described in \appref{app:training_details}; the all-pair case corresponds to $\mathcal{P}=\{(i,j):i\neq j\}$.
Here $c^{\mathrm{R}}$ and $c^{\mathrm{T}}$ weight the rotation and translation residuals separately, while the $-\log c$ terms prevent the model from assigning uniformly low confidence. This gives the model a way to downweight poorly constrained pairs during training and then reuse the same signal as aggregation weights and online keyframe-bank utilities at inference (and as edge weights in the optional pose-graph refinement of \appref{app:online_details}). The next diagnostic asks whether this training signal also behaves as geometric pair reliability.

\subsection{Learned Confidence as Pair Reliability}\label{app:confidence_error}

\begin{figure}[t]
\centering
\begin{subfigure}{0.49\linewidth}
  \centering
  \includegraphics[width=\linewidth]{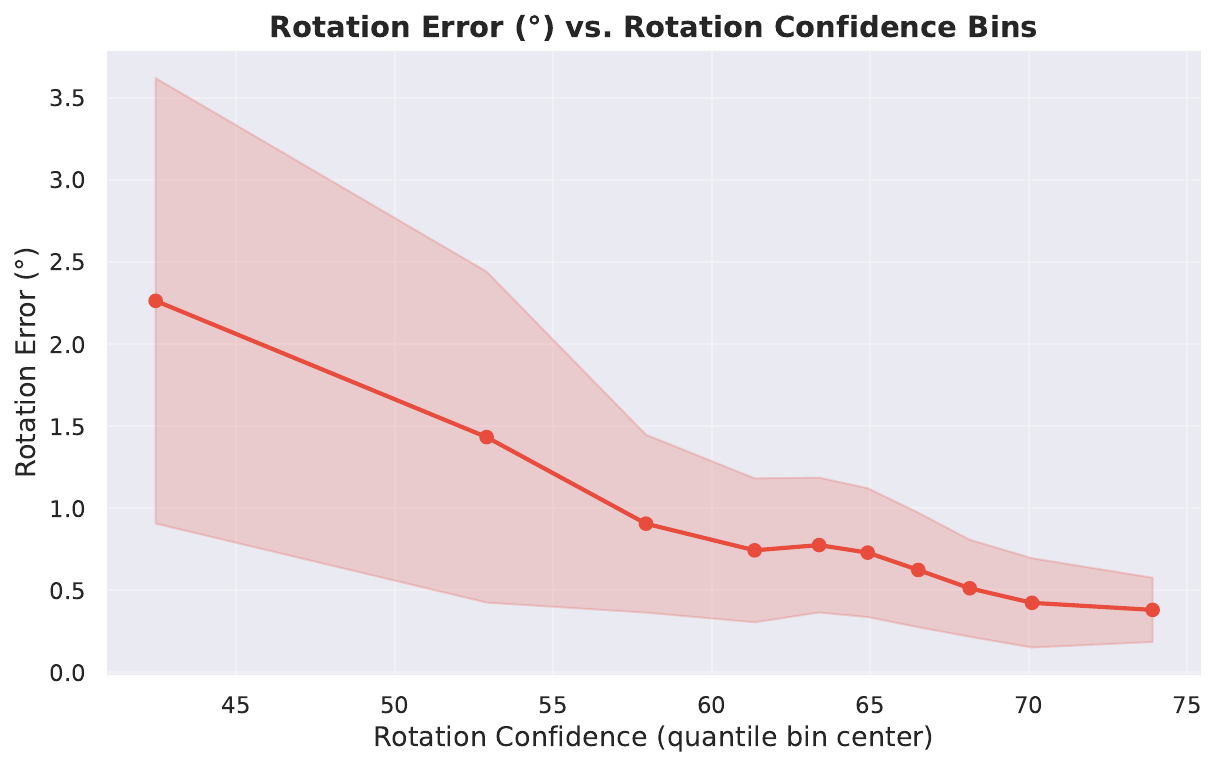}
  \caption{Rotation error binned by rotation confidence.}
  \label{fig:rot_error_conf}
\end{subfigure}
\hfill
\begin{subfigure}{0.49\linewidth}
  \centering
  \includegraphics[width=\linewidth]{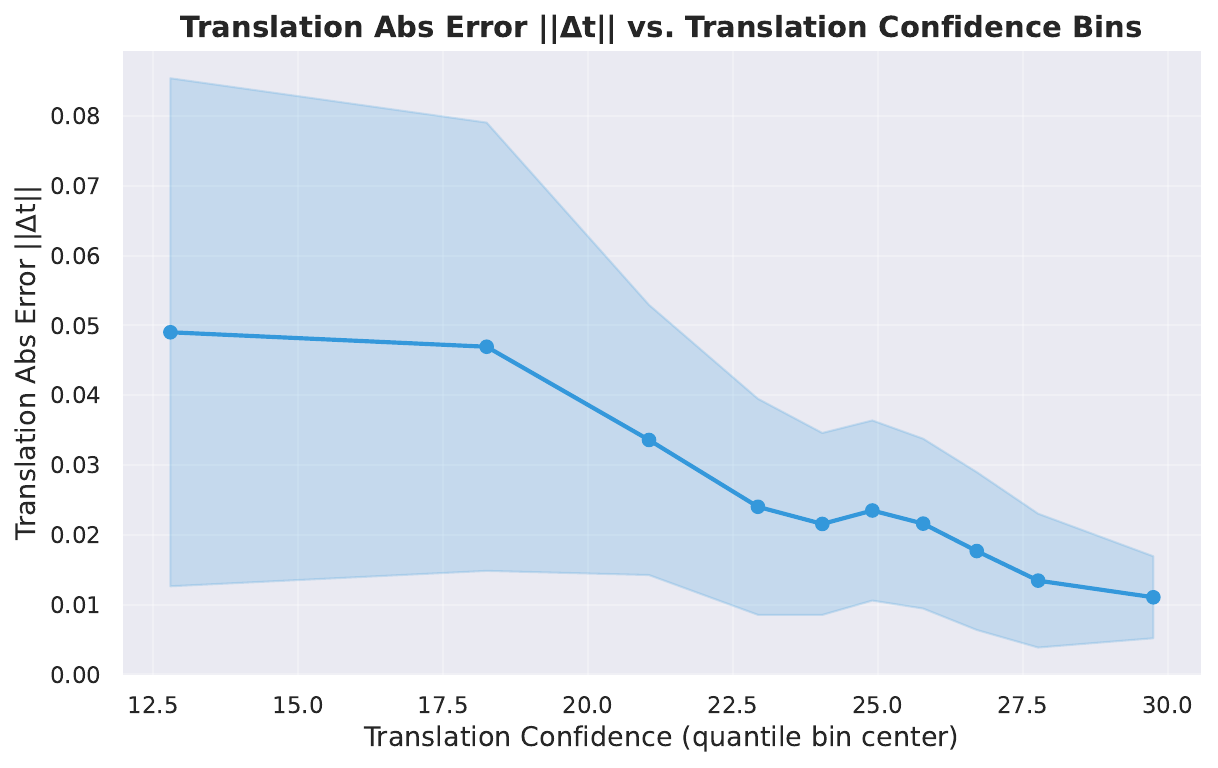}
  \caption{Translation error binned by translation confidence.}
  \label{fig:trans_error_conf}
\end{subfigure}
\caption{\textbf{Learned confidence behaves as pair reliability.} Across all pairs from the first $20$ ScanNet scenes, pairs are grouped into equal-mass confidence quantile bins; the $x$-axis is the bin center. The solid line shows the per-bin mean pose error and the shaded band shows the within-bin dispersion, so each polyline reports the average error and spread within that confidence quantile interval. Higher predicted confidence corresponds to lower mean error and tighter spread in the corresponding component.}
\label{fig:confidence_error_scannet}
\end{figure}

To check whether the predicted rotation and translation confidences act as learned pair-reliability estimates rather than only loss weights, we aggregate all image pairs from the first $20$ ScanNet test scenes~\cite{dai2017scannet} and bin them by the corresponding predicted confidence(\figref{fig:confidence_error_scannet}).

For both components, predicted confidence is monotone in the corresponding error: mean error decreases from low to high confidence bins, and the within-bin spread shrinks accordingly. The trend is sharpest at the low-confidence end, where unreliable pairs carry markedly larger residuals and wider variance. This is the property needed for the three downstream uses of confidence in our system: as a per-component loss weight in the Laplacian likelihood, as an edge weight in the PGO back-end, and as a frame-utility signal in the online keyframe bank.

The two panels also expose an asymmetry between components. Rotation confidence concentrates in a higher numerical range and reaches lower binned error than translation confidence, consistent with rotation being the easier learning target while translation remains more sensitive to baseline, scale, and parallax. A single shared pose confidence would have to average over this asymmetry, either over-trusting translation on rotation-easy pairs or under-using a reliable rotation edge when translation is ambiguous. Predicting the two confidences separately lets the model express this difference directly and lets each downstream consumer read the reliability of the component it actually depends on.

\section{Additional Ablations}\label{app:additional_ablations}

\paragraph{Full-attention fine-tuning reference.}
We compare a full-attention fine-tuned version of our model against the original DA3-Large~\cite{lin2025da3} backbone (\tabref{tab:full_attention_reference}). This is not a controlled ablation: DA3-Large is trained with substantially more data, whereas our variant uses the relative-pose objective from \secref{sec:relpose_vs_direct}. We therefore treat DA3-Large as a backbone calibration reference rather than a matched baseline.

\begin{table}[tbp]
\centering
\caption{\textbf{Full-attention fine-tuning reference.} Pose ATE (lower is better). DA3-Large is the original backbone trained with absolute-pose supervision and substantially more data; \modelname uses the relative-pose objective. The comparison is included as a calibration reference, not as a controlled ablation.}
\label{tab:full_attention_reference}
\small
\setlength{\tabcolsep}{6pt}
\begin{tabular}{l ccc}
\toprule
Variant & Sintel & TUM-dyn. & ScanNet \\
\midrule
DA3-Large~\cite{lin2025da3}                                 & 0.140 & 0.013 & 0.039 \\
\modelname (full-attention fine-tuned)                      & 0.117 & 0.011 & 0.035 \\
\bottomrule
\end{tabular}
\end{table}

The full-attention fine-tuned variant improves over the original backbone on these pose metrics despite using a relative-pose target rather than DA3's direct camera-pose supervision.

\subsection{Aggregation and Selection Heuristics}
\label{app:abl_aggregation}

We further investigate sensitivity to the aggregation strategy and the keyframe admission threshold $\tau$, which together govern the density and reliability of the reconstructed trajectory.

\begin{table}[t]
\centering
\caption{\textbf{Aggregation strategy} (left) and \textbf{acceptance threshold $\tau$} (right). Left: ATE on Sintel, TUM-dynamics, and ScanNet for top-$k$ confidence-weighted aggregation, all-pair averaging, and an optional pose-graph optimization (PGO) on top. Right: ScanNet ATE at $100$/$300$/$500$/$800$/$1000$ input frames for different $\tau$. \textbf{Bold}/underlined marks best/second-best per column.}
\label{tab:abl_aggregation}
\small
\setlength{\tabcolsep}{4pt}
\begin{minipage}[t]{0.48\linewidth}
\centering
\begin{tabular}{l cccc}
\toprule
Variant & TUM & Sintel & ScanNet & Avg.\ \\
\midrule
Top-1   & 0.0199 & 0.1272 & 0.0401 & 0.0624 \\
Top-5   & 0.0188 & 0.1177 & 0.0392 & 0.0586 \\
Top-10  & \third{0.0183} & \third{0.1153} & \third{0.0387} & \third{0.0574} \\
All-avg & \second{0.0179} & \second{0.1152} & \second{0.0384} & \second{0.0572} \\
+ PGO   & \best{0.0179} & \best{0.1142} & \best{0.0381} & \best{0.0567} \\
\bottomrule
\end{tabular}
\end{minipage}\hfill
\begin{minipage}[t]{0.48\linewidth}
\centering
\begin{tabular}{l ccccc}
\toprule
$\tau$ & 100 & 300 & 500 & 800 & 1k \\
\midrule
0.96 & \second{0.066} & 0.200 & 0.270 & 0.334 & 0.357 \\
0.97 & \best{0.061} & \third{0.180} & \third{0.250} & \third{0.306} & \third{0.332} \\
0.98 & \best{0.061} & \best{0.171} & \best{0.245} & \best{0.295} & \best{0.318} \\
0.99 & \third{0.068} & \second{0.172} & \second{0.246} & \second{0.300} & \second{0.329} \\
\bottomrule
\end{tabular}
\end{minipage}
\end{table}

\paragraph{Influence of neighborhood size $K$.}
We evaluate the impact of the number of reference frames $K$ used during pose fusion (\tabref{tab:abl_aggregation}, left). Accuracy improves as $K$ grows, and all-pair averaging gives the best closed-form ATE across the three benchmarks; top-$K$ aggregation is therefore best understood as an efficiency cap that trades a small amount of accuracy for lower fusion cost. Adding an optional confidence-weighted pose-graph optimization on top of the all-pair aggregate yields a further small but consistent improvement, indicating that the same confidence also serves as a useful per-pair weight in a global solve.

\paragraph{Keyframe admission threshold $\tau$.}
The threshold $\tau$ controls the feature novelty required for a frame to enter the keyframe bank. Since admission requires the maximum token similarity to fall below $\tau$, a lower threshold is more selective and a higher threshold is more permissive. Empirical results on ScanNet (\tabref{tab:abl_aggregation}, right) identify $\tau = 0.98$ as the best balance across varying sequence lengths: $\tau = 0.96$ can starve the bank in long-horizon streams, while $\tau = 0.99$ admits more redundant frames and noisy edges.

\section{Training Details}\label{app:training_details}

\paragraph{Architecture and trainable parameters.}
We build on the pretrained DA3-Large~\cite{lin2025da3} backbone and keep most of it frozen: only the global attention layers and the newly introduced pairwise pose head are updated, yielding roughly $110$M trainable parameters out of the $372$M-parameter inference model. The focal-length head is a small MLP that maps each camera token $\mathbf{z}_i$ to a single scalar $\hat{f}_i$ and is supervised by a plain $L_1$ residual against the ground-truth focal length. Training uses frame-causal global attention. The same checkpoint is used for both streaming and full-context inference; the latter removes the causal mask at test time so global attention runs bidirectionally over the full clip, then applies the lightweight pose-graph refinement described in \appref{app:online_details}.

\paragraph{Implementation details.}
The model is trained on sequences of $4$--$32$ views.
Most batches use clips of up to $16$ views, and the relative-pose stage extends the curriculum to $32$ views in the second half of training.
All training is done on $48$\,GB GPUs.
To fit long streams in this memory budget, we use gradient checkpointing, \texttt{bfloat16} mixed precision, and a FlexAttention~\cite{dong2024flexattention} implementation of frame-causal attention, which lets us pack about $200$ views per $48$\,GB GPU without the depth teacher and $96$ views per $48$\,GB GPU when the DA3 teacher is enabled.
For pose supervision, we evaluate the relative-pose head on the causal lower-triangular set of past-to-current ordered pairs after the DA3 forward pass. This provides $O(N^2)$ supervised relative edges during training without a comparable backbone cost, because the all-pair computation is only an MLP over compact camera tokens and is negligible relative to image-token attention. At inference time, streaming does not materialize all historical pairs: each incoming frame is paired only with the bounded active context $\mathcal{C}_t$, so per-frame cost is bounded by the keyframe bank.
Confidence logits are passed through a softplus nonlinearity to ensure positive confidence values.

\paragraph{Optimization schedule.}
The model is optimized with AdamW in two stages that share the same head and losses. \emph{Stage~1} adapts the DA3-Large backbone into a frame-causal absolute-pose checkpoint for $15$k iterations at a constant learning rate of $1\mathrm{e}{-4}$ with a dynamic batch size of $4$--$16$ views. \emph{Stage~2} trains the relative-pose head on top of this checkpoint for $25$k iterations, starting at $1\mathrm{e}{-4}$ and decaying with a cosine schedule to $1\mathrm{e}{-5}$, with gradient accumulation of $2$ and a dynamic batch size that begins at $4$--$16$ views and is extended to $4$--$32$ views in the second half of the schedule.

\paragraph{Training datasets.}
We train on a diverse collection of $15$ publicly available datasets spanning synthetic scenes, LiDAR/RGB-D captures, and COLMAP reconstructions. Table~\ref{tab:training_datasets} reports the number of training scenes or sequences retained after dataset-specific filtering. For AriaSyntheticENV, we use the first $2{,}000$ scenes, and for OmniWorld we use the OmniWorld-Game subset.

\begin{table}[t]
\centering
\caption{\textbf{Training datasets.} Counts denote the number of scenes or sequences used for training after dataset-specific filtering.}
\label{tab:training_datasets}
\small
\setlength{\tabcolsep}{6pt}
\begin{tabular}{l r l}
\toprule
Dataset & Count & Type \\
\midrule
AriaSyntheticENV~\cite{aria_ase} & 2,000 & Synthetic \\
ARKitScenes~\cite{baruch2021arkitscenes} & 4,388 & LiDAR/RGB-D \\
CO3Dv2~\cite{reizenstein2021co3d} & 30,616 & COLMAP \\
DL3DV~\cite{ling2024dl3dv} & 6,379 & COLMAP \\
HyperSim~\cite{roberts2021hypersim} & 344 & Synthetic \\
MapFree~\cite{arnold2022mapfree} & 921 & COLMAP \\
MVS-Synth~\cite{huang2018deepmvs} & 121 & Synthetic \\
OmniWorld-Game~\cite{zhou2025omniworld} & 484 & Synthetic \\
ScanNet~\cite{dai2017scannet} & 1201 & LiDAR/RGB-D \\
ScanNet++~\cite{yeshwanth2023scannetpp} & 230 & LiDAR/RGB-D \\
Spring~\cite{mehl2023spring} & 47 & Synthetic \\
TartanAir~\cite{wang2020tartanair} & 355 & Synthetic \\
Dynamic Replica~\cite{karaev2023dynamicreplica} & 524 & Synthetic \\
Virtual KITTI 2~\cite{cabon2020vkitti2} & 50 & Synthetic \\
WildRGBD~\cite{xia2024wildrgbd} & 23,050 & LiDAR/RGB-D \\
\bottomrule
\end{tabular}
\end{table}

\section{Robustness Details}\label{app:robustness_details}

\paragraph{Evaluation protocol.} We follow the Robust-VGGT~\cite{han2025emergent} controlled-distractor protocol with two modifications for streaming evaluation. (i)~We preserve the dataset's original frame order: distractor frames are interleaved into the clean stream rather than reshuffled, so \modelname is exposed to distractors at arbitrary points along the trajectory. (ii)~We keep the first three frames clean as a calibration prefix; this gives the streaming model a short, distractor-free initialization before the gating decision begins. For each trial we sample $N_c$ clean images from a single scene and $N_n$ distractor images drawn uniformly at random from other scenes in the same dataset, with disjoint pools and scene-agnostic selection. Three settings, \textbf{Small} / \textbf{Medium} / \textbf{Large}, differ only in $N_n$. On RobustNeRF~\cite{sabour2023robustnerf} we use $N_c{=}30$ and $N_n{\in}\{10,30,50\}$. ETH3D~\cite{schops2017eth3d} has fewer frames per scene, so we use $N_c{=}14$ and $N_n{\in}\{5,14,30\}$. Each setting is repeated for $10$ trials with different seeds and we report the mean. All methods are evaluated on identical sampled sets, and the same protocol is used for both the pose and depth/point-map evaluations.

\paragraph{Balanced filtering score.} The main robustness table reports the distractor rejection success rate, matching Robust-VGGT. To guard against the degenerate strategy of rejecting too many frames, we additionally compute a balanced filtering score,
\[
\mathrm{BFS}=\tfrac{1}{2}\,\mathrm{DistractorReject}+\tfrac{1}{2}\,\mathrm{CleanAccept}.
\]
Under the Small/Medium/Large settings of \tabref{tab:robustness}, the \modelname confidence gate obtains balanced filtering scores of $0.84/0.89/0.93$ on ETH3D and $0.999/0.992/0.996$ on RobustNeRF, indicating that high distractor rejection does not come from uniformly rejecting clean frames.

\section{Long-Sequence Pose Evaluation on DL3DV-Benchmark}\label{app:dl3dv_pose}

We provide the protocol details for the DL3DV-Benchmark~\cite{ling2024dl3dv} evaluation reported in \tabref{tab:dl3dv_pose} of the main paper. The evaluation is purely camera-trajectory based. We report Absolute Trajectory Error (ATE) after Sim(3) alignment, together with per-trajectory rotation RMSE. Aggregates are mean values over a random subset of $25$ scenes drawn from the $140$ DL3DV-Benchmark scenes, with per-scene sequence lengths ranging from $304$ to $439$ frames. All methods process every frame in the sequence. \modelname runs the full sequence in a \emph{single pass with no reset}; in our DL3DV run, TTT3R~\cite{chen2025ttt3r} uses a $100$-frame reset interval, which gives better results than no-reset inference.

The reported numbers are consistent with the long-sequence pose curves in \secref{sec:longseq}: the keyframe-bank front-end with confidence-weighted relative-pose aggregation remains stable at the horizons covered by DL3DV-Benchmark, where recurrent and KV-cache baselines accumulate visible drift.

\section{Video Depth Estimation}\label{app:video_depth}

We additionally evaluate video depth estimation against the DA3-Large backbone on Sintel~\cite{butler2012sintel}, Bonn~\cite{palazzolo2019refusion}, and KITTI~\cite{geiger2012kitti}. Following the standard protocol, we report Absolute Relative error (Abs Rel, lower is better) and $\delta<1.25$ accuracy (higher is better) under per-sequence median scale alignment.

\begin{table}[tbp]
\centering
\caption{\textbf{Video depth estimation on Sintel, Bonn, and KITTI.} Abs Rel is lower-is-better; $\delta<1.25$ is higher-is-better. \textbf{Bold} marks the best result per column.}
\label{tab:video_depth}
\small
\setlength{\tabcolsep}{5pt}
\begin{tabular}{l cc cc cc}
\toprule
& \multicolumn{2}{c}{Sintel} & \multicolumn{2}{c}{Bonn} & \multicolumn{2}{c}{KITTI} \\
\cmidrule(lr){2-3} \cmidrule(lr){4-5} \cmidrule(lr){6-7}
Method & Abs Rel$\downarrow$ & $\delta\!<\!1.25$$\uparrow$ & Abs Rel$\downarrow$ & $\delta\!<\!1.25$$\uparrow$ & Abs Rel$\downarrow$ & $\delta\!<\!1.25$$\uparrow$ \\
\midrule
DA3-Large~\cite{lin2025da3} & 0.460 & \textbf{0.518} & 0.106 & \textbf{0.890} & \textbf{0.122} & \textbf{0.870} \\
\modelnamebf                & \textbf{0.410} & 0.479 & \textbf{0.102} & 0.880 & 0.130 & 0.845 \\
\bottomrule
\end{tabular}
\end{table}

Our model matches the DA3-Large teacher on Bonn, improves Abs Rel on Sintel, and trails slightly on KITTI, showing that joint relative-pose and depth training preserves the backbone's depth quality despite the added pose objective.

\section{Limitations and Future Work}\label{app:limitations}

\begin{figure}[H]
  \centering
  \begin{subfigure}[b]{0.36\linewidth}
    \centering
    \includegraphics[height=4.2cm,keepaspectratio]{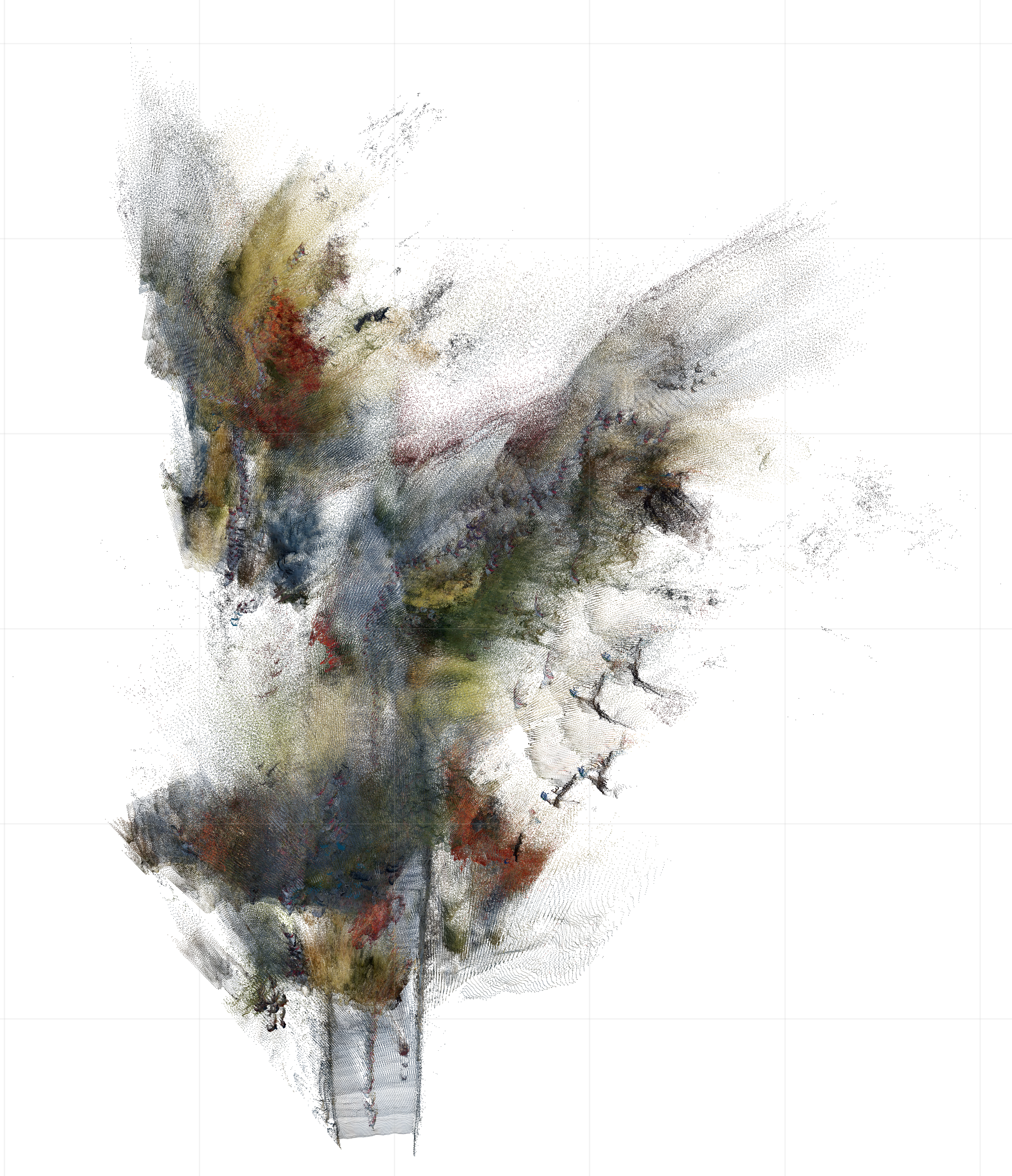}
    \caption{Without reset.}
  \end{subfigure}
  \hfill
  \begin{subfigure}[b]{0.55\linewidth}
    \centering
    \includegraphics[height=4.2cm,keepaspectratio]{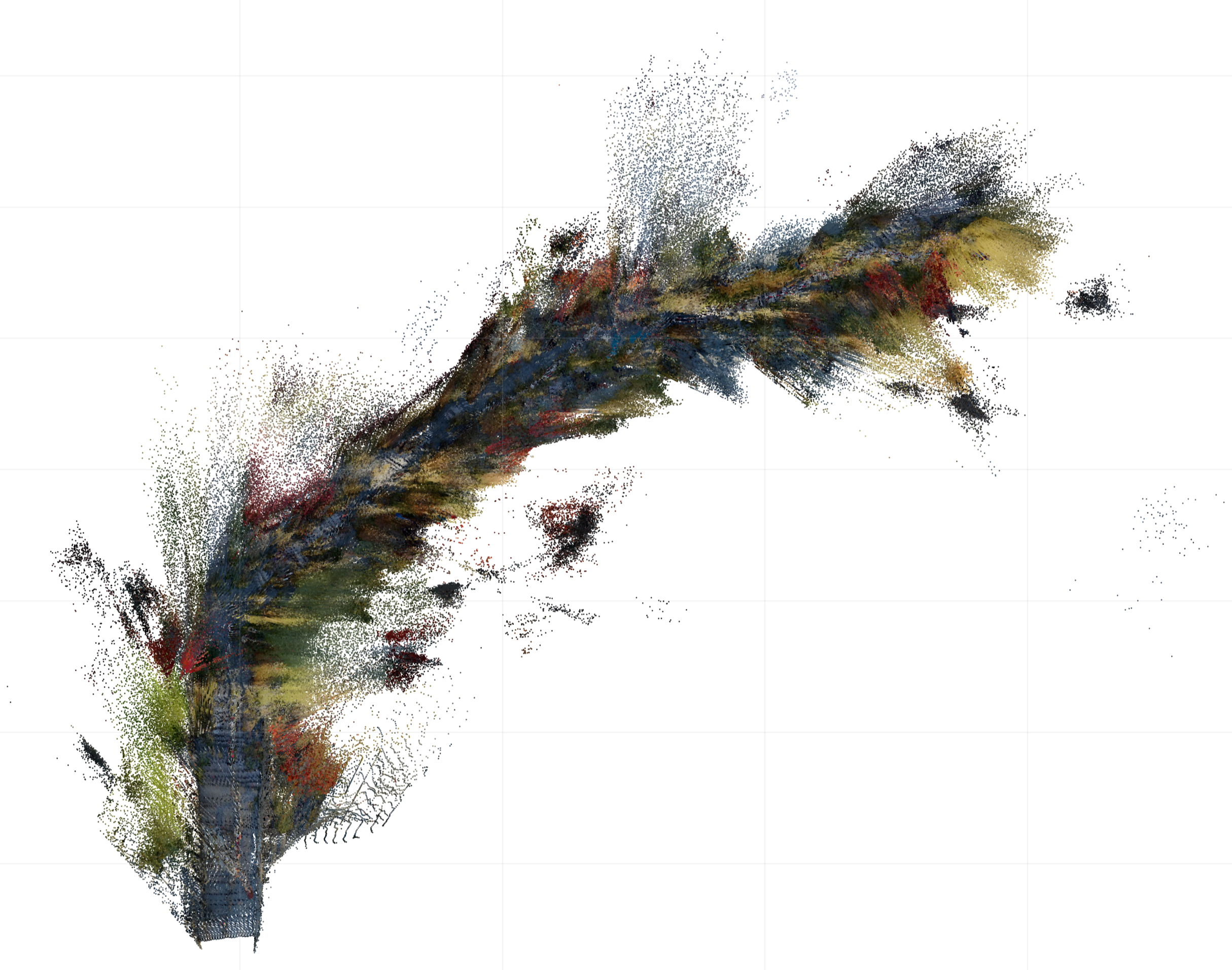}
    \caption{With periodic reset.}
  \end{subfigure}
  \caption{Bird's-eye view of a long streaming sequence. Without reset, the trajectory eventually drifts and the camera is lost; a periodic reset restores a clean track.}
  \label{fig:limit_bev}
\end{figure}

\begin{figure}[H]
  \centering
  \begin{subfigure}[t]{0.48\linewidth}
    \centering
    \includegraphics[width=\linewidth]{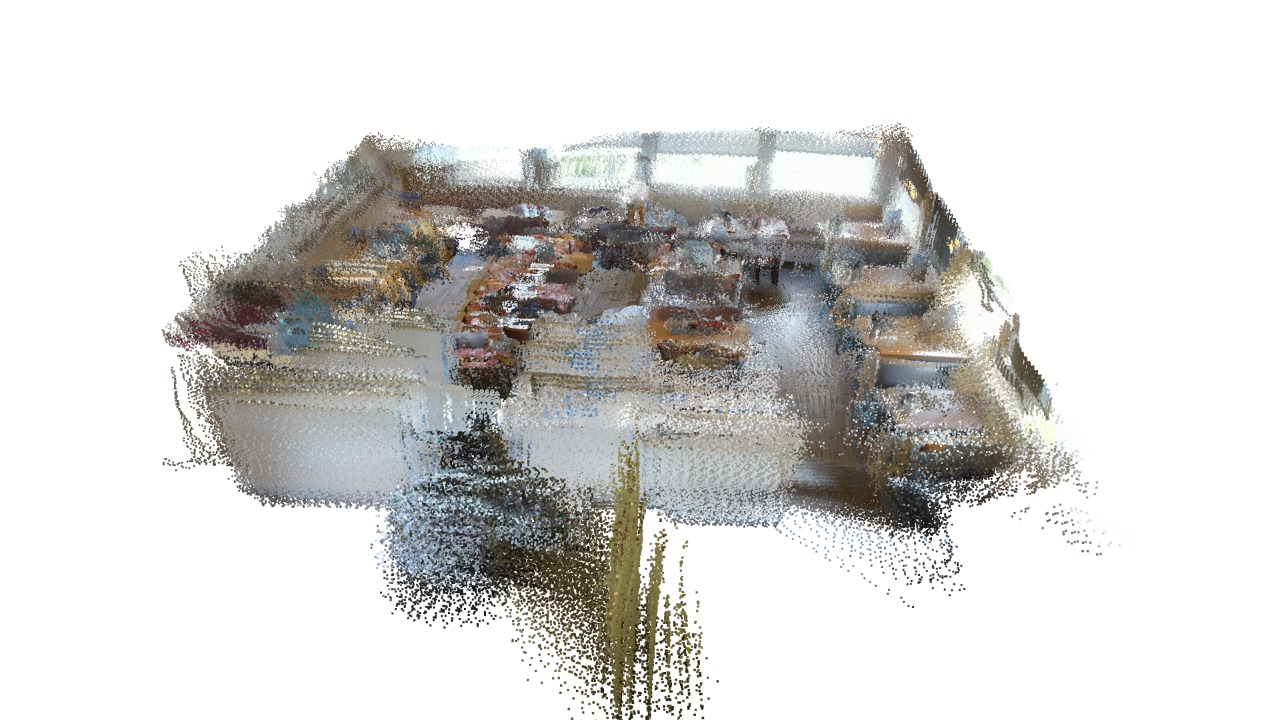}
  \end{subfigure}
  \hfill
  \begin{subfigure}[t]{0.48\linewidth}
    \centering
    \includegraphics[width=\linewidth]{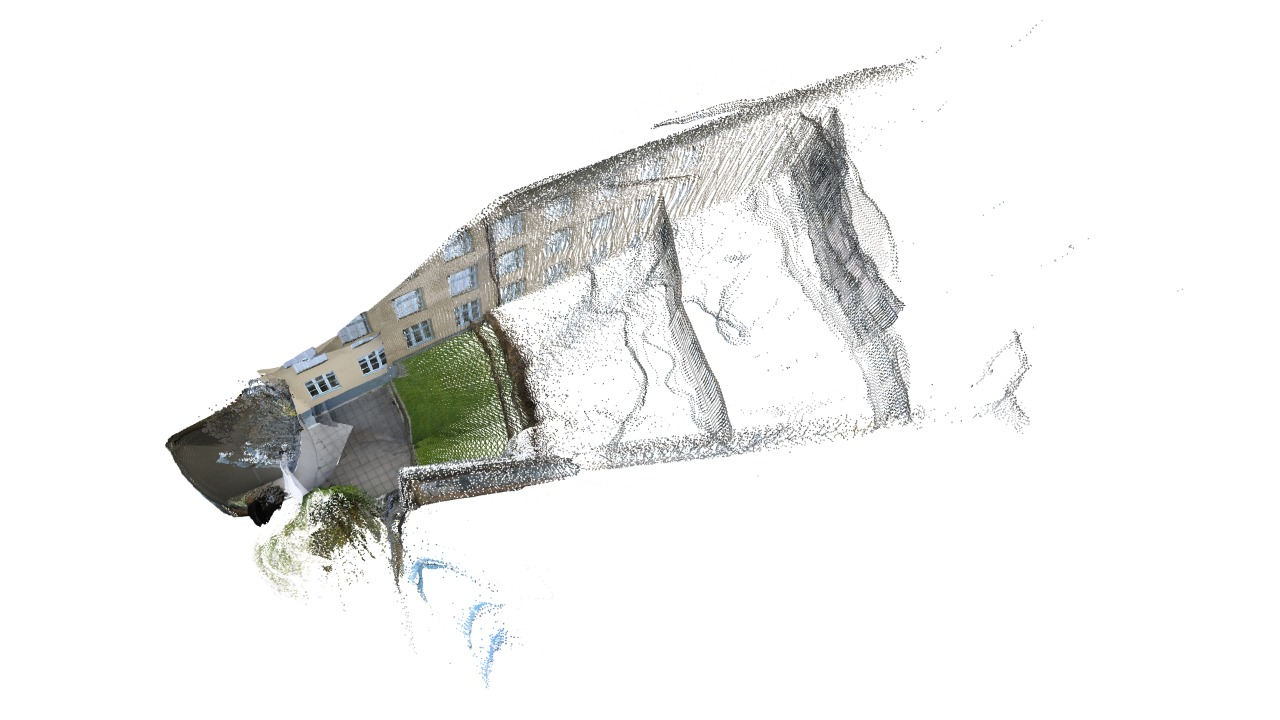}
  \end{subfigure}
  \caption{Causal streaming reconstructions. Distant viewpoints, occlusions, and long sequences can leave residual inconsistencies in the recovered geometry.}
  \label{fig:limit_viewer}
\end{figure}

\modelname is a 372M-parameter model trained on six 48\,GB GPUs, and we have not explored larger backbones or substantially more data. The keyframe bank still relies on a few hand-set thresholds ($\tau$, $\Delta_{\max}$, $M_{\max}$, outlier-gate constants) that generalize across our benchmarks but may need re-tuning on very different domains. On very long streams \modelstream can still drift without a periodic reset (\figref{fig:limit_bev}), and because it cannot revisit past tokens, distant viewpoints, occlusions, or long sequences can leave residual inconsistencies in the geometry (\figref{fig:limit_viewer}).

Natural next steps include scaling the backbone and training data, replacing the lightweight pose-only refinement with bundle adjustment or joint depth--pose optimization, learning the keyframe-bank thresholds and a reset policy in place of the current scalars, and a global refinement or limited bidirectional re-attention pass to address causal inconsistencies in the streaming setting.

\end{document}